\documentclass{article}

\usepackage{microtype}
\usepackage{graphicx}
\usepackage{subcaption}
\usepackage{booktabs} 

\usepackage{hyperref}

\usepackage[preprint]{icml2026}

\usepackage{amsmath}
\usepackage{amssymb}
\usepackage{mathtools}
\usepackage{amsthm}
\usepackage{bm}
\usepackage[capitalize,noabbrev]{cleveref}

\theoremstyle{plain}
\newtheorem{theorem}{Theorem}[section]

\theoremstyle{definition}
\newtheorem{definition}[theorem]{Definition}

\theoremstyle{remark}

\usepackage[textsize=tiny]{todonotes}

\icmltitlerunning{Submission and Formatting Instructions for ICML 2026}

\begin{document}

\twocolumn[
  \icmltitle{GeMA: Learning Latent Manifold Frontiers\\
     for Benchmarking Complex Systems}



  \icmlsetsymbol{equal}{*}

  \begin{icmlauthorlist}
    \icmlauthor{Jia Ming Li}{}
    \icmlauthor{Anupriya}{}
    \icmlauthor{Daniel J. Graham}{}
  \end{icmlauthorlist}
  

  \icmlcorrespondingauthor{Jia Ming (Simon) Li}{jiaming.li@imperial.ac.uk}

  \icmlkeywords{Latent variable models, Deep generative models, Manifold learning, Frontier estimation, Efficiency analysis}

  \vskip 0.3in
]



\printAffiliationsAndNotice{Transport Strategy Centre, Imperial College London, London, SW7 2AZ, United Kingdom} 

\newcommand{\theframe}{\textsc{GeMA}}
\newcommand{\themodel}{\textsc{ProMan-VAE}}

\begin{abstract}
Benchmarking the performance of complex systems such as urban and national rail networks, renewable generation assets and national economies is an important component of transport planning, regulation and macroeconomic analysis. Classical frontier methods, in particular Data Envelopment Analysis (DEA) and Stochastic Frontier Analysis (SFA), estimate an efficient frontier in the observed input--output space and define efficiency as distance to this frontier. While empirically useful, these approaches often rely on restrictive assumptions on the production possibility set, convexity, separability or specific parametric functional forms, and address structural heterogeneity and scale effects only indirectly.

We propose \textit{Geometric Manifold Analysis} ($\theframe$), a latent manifold frontier framework implemented via a productivity-manifold variational autoencoder (\themodel). Instead of specifying a frontier function in the observed space, $\theframe$ represents the production set as the boundary of a low-dimensional manifold embedded in the joint input--output space. A split-head encoder learns disentangled latent variables that capture technological structure and operational inefficiency. Efficiency is evaluated with respect to the learned manifold; endogenous peer groups arise as clusters in latent technology space; a quotient construction supports scale-invariant benchmarking; and a local certification radius, derived from the decoder Jacobian and a Lipschitz bound, quantifies the geometric robustness of individual efficiency scores.

We validate the framework on synthetic data designed to exhibit non-convex frontiers, heterogeneous technologies and scale bias, and on four real-world case studies: global urban rail systems (COMET), British rail operators (ORR), national economies (Penn World Table) and a high-frequency wind farm dataset (WF). Across these domains $\theframe$  behaves comparably to established methods when classical assumptions are approximately satisfied, and it appears to provide additional insight in settings with pronounced heterogeneity, non-convexity or size-related bias. We conclude by outlining how the static framework may be extended to a dynamic latent state-space ``world model'' of complex production systems, opening a path towards counterfactual analysis and policy design on the learned manifold.
\end{abstract}


\section{Introduction}
\label{sec:intro}

Benchmarking the performance of complex systems is a widely used instrument in governance, regulation and strategic planning. Urban and national rail networks, renewable energy assets and national economies are routinely compared using efficiency scores derived from frontier methods. Conceptually, the task is to estimate an efficient frontier describing the best attainable outputs for given inputs, and to quantify how far each decision-making unit (DMU) lies from this frontier.

Two main families of methods have been particularly influential in this field. Stochastic Frontier Analysis (SFA) specifies a parametric production function with a composed error term that separates noise and inefficiency \citep{aigner1977formulation, meeusen1977efficiency, Greene1993, Sickles2019}. Data Envelopment Analysis (DEA) constructs a piecewise-linear efficient frontier enveloping the data via linear programming \citep{charnes1978measuring, coelli2005introduction}. Both approaches have been widely applied in transport, energy, health and macroeconomic applications \citep[e.g.][]{oum1999survey, graham2008productivity, simar2015statistical} and have become standard tools in regulatory benchmarking \citep{smith2005role, rostamzadeh2021application}.

However, contemporary datasets pose several structural challenges. First, \emph{heterogeneity}: DMUs may operate under markedly different technological or organisational regimes, so that a single global frontier is difficult to interpret and may systematically favour some groups over others. Secondly, \emph{non-convexity and non-linearity}: indivisible investments, network effects and physical constraints can lead to production sets that are not well approximated by convex hulls or simple parametric forms \citep{keshvari2013stochastic}. Thirdly, \emph{scale bias}: in macroeconomic and infrastructure datasets, larger entities often receive higher efficiency scores simply because of their absolute size, even after controlling for standard inputs. Finally, there is a \emph{trust} question: efficiency scores are commonly reported as point estimates, with limited information on their stability to small perturbations in data or model specification.

Recent work has sought to relax some of the structural assumptions in classical frontier analysis. Convex non-parametric least squares (CNLS) and related approaches reinterpret DEA as a convex regression problem and impose shape constraints via inequalities on regression coefficients \citep{daraio2006robust, kuosmanen2010data}. Stochastic non-convex envelopment and order-$m$ frontiers explore non-convex production sets and robustness to extreme points \citep{simar2007improve, keshvari2013stochastic}. In parallel, machine learning methods have been used to estimate production functions with greater flexibility, employing kernel methods, tree ensembles and neural networks \citep{Breiman2001_RF, chen2016xgboost, Goodfellow-et-al-2016}. Deep learning has also been brought into efficiency analysis, both for predicting outputs and for re-evaluating DEA benchmarks \citep{bose2015neuraldea, guerrero2022combining, tsionas2023bayesian}. These developments, whilst valuable, typically retain the view of the frontier as a function in the observed input--output space and do not fully exploit the potential of latent variable modelling.

In this paper, we take a complementary perspective. Building on the manifold hypothesis in representation learning \citep{bengio2013representation} and ideas from geometric deep learning \citep{bronstein2017geometric, bronstein2021geometric}, we model the production possibility set as a low-dimensional, non-linear manifold embedded in the joint input--output space, and treat the frontier as the Pareto boundary of the production set induced by a latent variable model. We introduce \emph{Geometric Manifold Analysis} (\theframe), implemented by a productivity-manifold variational autoencoder ($\themodel$), which learns a latent technology coordinate and a separate inefficiency factor from data. This construction enables us to represent flexible, potentially non-convex frontiers, accommodate heterogeneous technologies as distinct regions of a shared manifold, define a quotient manifold that reduces certain scale effects, and attach a simple geometric robustness score to each efficiency estimate.

Building on these ideas, our contributions are as follows:
\begin{itemize}
    \item We introduce \emph{Geometric Manifold Analysis} (\theframe), a latent-manifold frontier framework in which the production possibility set is defined as the image of a low-dimensional manifold in joint input--output space. The associated $\themodel${} model disentangles latent technology from inefficiency under basic economic shape constraints, yielding a generative SFA-style formulation with explicit production-set semantics.
    \item We propose two geometric diagnostics for efficiency analysis: (i) a quotient construction that supports scale-invariant benchmarking by factoring out joint rescalings of inputs and outputs, and (ii) a local certification radius derived from the decoder Jacobian, which provides an interpretable indicator of the local robustness of efficiency scores.
    \item We empirically study $\theframe$ on synthetic data and three complex domains: national rail operators (ORR), a high-frequency wind farm dataset (WF) and macroeconomic data (PWT), with an additional urban rail case (COMET) in the appendix. The experiments show that $\theframe$ behaves comparably to classical frontier estimators when their assumptions are approximately satisfied, and can offer additional insight in settings with non-convex technologies, unobserved heterogeneity or scale-related bias.
\end{itemize}

The remainder of the paper is organised as follows. Section~\ref{sec:relatedwork} briefly reviews related work in frontier analysis, latent variable models and geometric deep learning. Section~\ref{sec:method} introduces the $\theframe$ framework and the $\themodel$ architecture. Section~\ref{sec:experiments} presents synthetic experiments and case studies on wind farms, national rail operators and macroeconomic data. Section~\ref{sec:discussion} discusses implications and limitations, and Section~\ref{sec:conclusion} concludes. Detailed derivations, additional experiments and data descriptions are provided in the appendices.

\section{Related Work}
\label{sec:relatedwork}

\paragraph{Frontier and efficiency analysis.}
Classical efficiency analysis builds on the notion of a production set and its efficient frontier \citep{farrell1957measurement}. Stochastic Frontier Analysis (SFA) specifies a parametric production function, often Cobb--Douglas or Translog, with a composed error term that separates statistical noise from a one-sided inefficiency component \citep{aigner1977formulation, meeusen1977efficiency, Greene1993, Sickles2019}. This econometric approach supports statistical inference and panel extensions, but is sensitive to functional-form misspecification and typically relies on restrictive distributional assumptions. Data Envelopment Analysis (DEA) and related non-parametric methods construct a piecewise-linear frontier that envelops the data using linear programming \citep{charnes1978measuring, coelli2005introduction}. Extensions relax disposability and returns-to-scale assumptions or consider non-convex hulls and order-$m$ frontiers \citep{deprins1984measuring, simar2007improve, keshvari2013stochastic}. These methods have been widely applied in transport, energy and other infrastructure sectors \citep[e.g.][]{oum1999survey, graham2008productivity, rostamzadeh2021application}, but typically operate in the observed input--output space and impose convexity or parametric structure on the production set.

\paragraph{Machine learning and deep latent variable models.}
Machine learning methods have increasingly been used to estimate production functions and efficiency scores with greater flexibility, employing kernel methods, tree ensembles and neural networks \citep{Breiman2001_RF,chen2016xgboost, Goodfellow-et-al-2016}. Recent work has explored the use of deep learning architectures to model complex production relationships and inefficiency \citep{bose2015neuraldea, guerrero2022combining, tsionas2023bayesian}. Deep generative models and variational autoencoders (VAEs) introduce explicit latent variables to capture low-dimensional structure underlying high-dimensional observations \citep{kingma2013auto, rezende2014stochastic}, with disentangled representations aiming to separate latent factors with distinct semantic roles \citep{bengio2013representation, higgins2017beta}. In the context of efficiency analysis, most deep learning approaches focus on prediction quality or on re-scoring DEA benchmarks, and do not explicitly define a production set or frontier in latent space. In contrast, $\theframe$ uses a VAE-style architecture to define a generative frontier model with explicit production-set semantics and an inefficiency factor.

\paragraph{Geometric and structured deep learning.}
Geometric deep learning emphasises the importance of exploiting the underlying geometric structure of data---manifolds, graphs and more general structured domains \citep{bronstein2017geometric, bronstein2021geometric}. The manifold hypothesis suggests that high-dimensional observations often concentrate near lower-dimensional manifolds; methods such as Isomap and locally linear embedding (LLE) \citep{roweis2000nonlinear,tenenbaum2000global} and more recent approaches such as UMAP \citep{mcinnes2018umap} provide tools for learning or visualising manifold structure. In parallel, latent manifold models have been proposed in scientific domains to interpret complex data geometry \citep{lopez2018deep, moon2019visualizing, nieh2021geometry, perich2025neural}. In econometrics and operations research, geometric ideas have begun to appear in the analysis of efficient frontiers and multi-objective optimisation \citep{chatigny2024learning, felten2024toolkit}, and there is growing interest in causal representation learning on learned manifolds \citep{scholkopf2021toward}. $\theframe$ draws inspiration from this geometric perspective but uses relatively standard encoder--decoder architectures; the geometric structure enters through the interpretation of the learned latent space as a productivity manifold and through the quotient and certification constructions that we use for benchmarking and robustness diagnostics.

\section{GeMA: Latent Manifold Frontiers}
\label{sec:method}

We now formalise \emph{Geometric Manifold Analysis} (\theframe) and its $\themodel$ implementation. We first define a productivity manifold and the induced production set, then specify a generative model that disentangles latent technology from inefficiency. We subsequently introduce two geometric diagnostics: a quotient construction for scale-invariant benchmarking and a local certification radius for robustness assessment.

\subsection{Productivity manifold and production set}
\label{subsec:prod_set}

Let $\mathcal{X} \subset \mathbb{R}^d$ denote the input space and $\mathcal{Y} \subset \mathbb{R}^v$ the output space. Classical production theory assumes the existence of a production set $\mathcal{T} \subset \mathcal{X} \times \mathcal{Y}$ such that $(x,y) \in \mathcal{T}$ if and only if output $y$ is feasible with inputs $x$. The output-oriented efficient frontier is the Pareto-efficient boundary
\[
\partial \mathcal{T} = \{(x,y) \in \mathcal{T}: \nexists \, y' \ge y \text{ with } (x,y') \in \mathcal{T}\},
\]
with the inequality interpreted component-wise.

$\theframe$ defines the production set constructively via a low-dimensional manifold embedded in the joint input--output space. Let $z \in \mathbb{R}^K$ denote a latent technology coordinate and let
\[
g_\theta : \mathcal{X} \times \mathbb{R}^K \to \mathbb{R}^v
\]
be a decoder network with parameters $\theta$. The \emph{productivity manifold} is
\[
\mathcal{M}_\theta \;=\; \{ (x, y) \in \mathcal{X} \times \mathbb{R}^v : \exists z \in \mathbb{R}^K \text{ with } y = g_\theta(x,z)\}.
\]
For each input--technology pair $(x,z)$ the decoder produces a point $(x, g_\theta(x,z))$ on the manifold. The production set induced by $g_\theta$ is then defined as
\[
\mathcal{T}_\theta \;=\; \{ (x,y) \in \mathcal{X} \times \mathbb{R}^v : \exists z \in \mathbb{R}^K \text{ with } y \le g_\theta(x,z)\},
\]
with $y \le g_\theta(x,z)$ interpreted component-wise. This captures the idea that any output vector not exceeding a frontier point in each component is feasible. The estimated frontier is the Pareto-efficient boundary $\partial \mathcal{T}_\theta$.

The latent coordinate $z$ may be viewed as a set of intrinsic structural parameters describing technology or operating conditions. Different regions of latent space encode different technological regimes or business models, and efficiency is assessed relative to the geometry of $\mathcal{T}_\theta$ at a given $(x,z)$.

\subsection{$\themodel$ generative model}
\label{subsec:proman}

We observe a dataset $\{(x_i,y_i)\}_{i=1}^N$ of decision-making units (DMUs). The $\themodel$ model treats each observed input--output pair as a noisy realisation of a two-stage process. In the first stage, a unit adopts a structural paradigm or \emph{technology} $z_i$, which locates it on the productivity manifold $\mathcal{M}_\theta$. In the second stage, the corresponding frontier output is scaled down by an \emph{inefficiency} factor and contaminated by random noise. The model is trained to invert this process: given an observation $(x_i,y_i)$, it infers a latent technology $z_i$ and an inefficiency $u_i$ that best explain the data.

Concretely, we introduce latent variables $z_i \in \mathbb{R}^K, u_i \in [0,\infty)$,
where $z_i$ represents technology and $u_i$ represents inefficiency. We place a standard normal prior on technology and an exponential prior on inefficiency, $z_i \sim \mathcal{N}(0, I), u_i \sim \mathrm{Exp}(\lambda).$
Given $(x_i,z_i)$, the decoder network $g_\theta$ produces a (theoretical) frontier output $y_i^\ast = g_\theta(x_i,z_i).$
Actual output is modelled as
\[
y_i = y_i^\ast \exp(-u_i) \varepsilon_i,
\]
where $\varepsilon_i$ is a multiplicative noise term. In practice we work in log-space and approximate the noise as additive Gaussian, yielding an SFA-style structural equation with the parametric frontier $f$ replaced by the manifold mapping $g_\theta(x,z)$. Full details of the log-space likelihood and variational objective are given in Appendix~A.

\begin{figure}[htbp]
    \centering
    \includegraphics[width=\columnwidth]{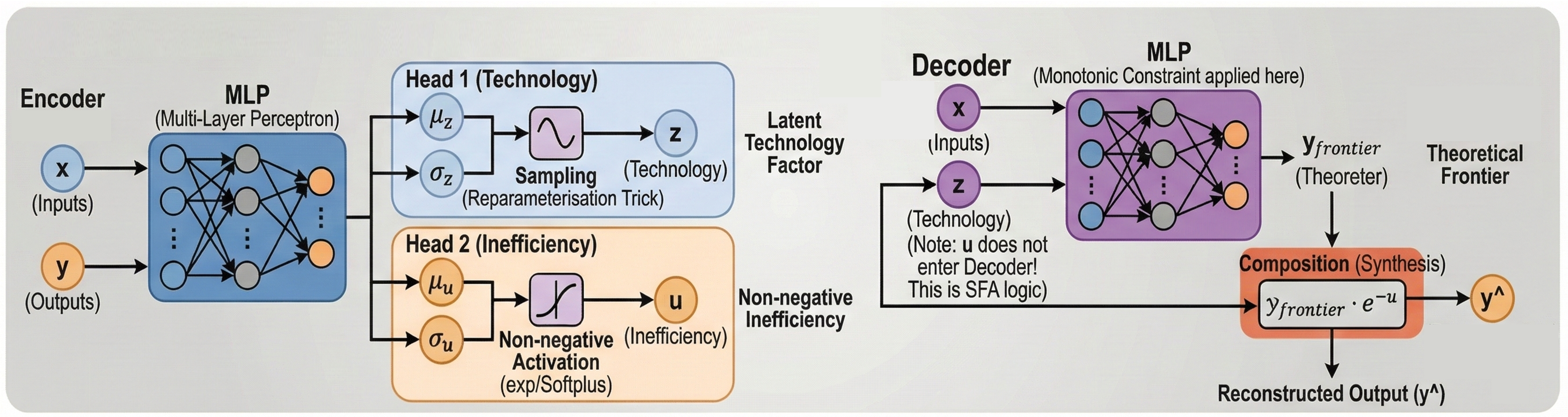}
    \caption{\textbf{$\themodel$ architecture.} A split-head encoder maps observed inputs and outputs $(\mathbf{x},\mathbf{y})$ to latent technology $(\mathbf{z})$ and inefficiency $(u)$. The decoder $\mathcal{G}_\theta$ reconstructs the frontier output from $(\mathbf{x},\mathbf{z})$, and the realised output is obtained by scaling the frontier with $\exp(-u)$. A monotonicity regulariser encourages the decoder to be weakly increasing in each input dimension.}
    \label{fig:architecture}
\end{figure}

\paragraph{Encoder and latent disentanglement.}
To infer $(z_i,u_i)$ from data, $\themodel$ employs a split-head encoder $q_\phi(z_i,u_i \mid x_i,y_i)$ with parameters $\phi$. A shared multilayer perceptron processes the concatenated inputs and outputs and branches into two heads:
\begin{itemize}
    \item a \emph{technology head} that outputs a mean vector $\mu^{(z)}_i \in \mathbb{R}^K$ and a log-variance vector $\log \sigma^{2,(z)}_i \in \mathbb{R}^K$, defining a diagonal Gaussian posterior $q_\phi(z_i \mid x_i,y_i) = \mathcal{N}(\mu^{(z)}_i, \operatorname{diag}(\sigma^{2,(z)}_i))$;
    \item an \emph{inefficiency head} that outputs parameters of a non-negative distribution for $u_i$, for example by specifying a Gaussian posterior for $\log u_i$ and mapping it through an exponential or softplus transformation, so that $u_i \ge 0$.
\end{itemize}
Sampling from $q_\phi(z_i,u_i \mid x_i,y_i)$ is implemented via the reparameterisation trick, allowing gradients to propagate through stochastic nodes during training.

To respect basic economic logic, we impose an approximate monotonicity constraint on $g_\theta$ by adding a penalty term $\mathcal{R}_{\mathrm{mono}}(\theta)$ to the loss, encouraging non-negative marginal products in each input dimension. In practice, we estimate partial derivatives of $g_\theta$ with respect to inputs by finite differences over a grid of reference points and penalise negative increments. Further details on the network architecture and regularisation are provided in Appendix~B.

\paragraph{Variational objective and efficiency scores.}
The model is trained by maximising an evidence lower bound (ELBO) on the log-likelihood of the outputs given the inputs. For a single observation $(x_i,y_i)$ the ELBO has the generic form
\begin{equation*}
\begin{split}
\mathcal{L}_i(\theta,\phi)
= \mathbb{E}_{q_\phi(z_i,u_i \mid x_i,y_i)}\big[\log p_\theta(y_i \mid x_i,z_i,u_i)\big]\\
- \mathrm{KL}\big(q_\phi(z_i,u_i \mid x_i,y_i) \,\Vert\, p(z_i,u_i)\big),
\end{split} 
\end{equation*}
where $p(z_i,u_i) = p(z_i)p(u_i)$ is the prior and $p_\theta(y_i \mid x_i,z_i,u_i)$ is induced by the log-space structural equation. In our implementation, closed-form expressions are available for the Kullback--Leibler terms, and the reconstruction term is approximated by a per-output Huber or squared loss on log-outputs. Aggregating over the dataset and adding the monotonicity penalty yields the overall training objective. Derivations and implementation details are given in Appendix~A.

Under this model, the inefficiency variable $u_i$ provides a natural scalar efficiency measure: the factor $\exp(-u_i)$ scales the frontier output downwards, so we define an output-oriented efficiency index $\mathrm{Eff}_i = \exp(-u_i) \in (0,1]$. In principle, one may also consider distance-based measures defined via the geometry of $\mathcal{T}_\theta$, such as the minimal distance in output space between $y_i$ and the frontier at input $x_i$. In this work we primarily use $\exp(-u_i)$, as it is directly learned by the model and has a simple interpretation as the proportion of frontier output.

The latent technology vectors $z_i$ provide a low-dimensional representation of structural heterogeneity. After training, we may project the $z_i$ into two or three dimensions using a manifold visualisation method such as UMAP and cluster them to identify endogenous peer groups. These clusters can be interpreted as regions of the latent manifold corresponding to distinct technological regimes or business models.

\subsection{Quotient manifold for scale-invariant benchmarking}
\label{subsec:quotient}

In macroeconomic and infrastructure applications, absolute size (for example measured by GDP, population or network length) can strongly influence efficiency scores. Larger countries or systems may appear efficient simply because they operate at a greater scale, even if their underlying technology is similar to that of smaller units. To address this, we introduce a simple equivalence relation that identifies scale-equivalent production points.

Given two points $(x,y)$ and $(x',y')$ in $\mathcal{T}_\theta$, we write
\[
(x,y) \sim (x',y') \quad \text{if} \quad \exists \lambda > 0 \text{ such that } (x',y') = (\lambda x, \lambda y).
\]
Intuitively, two configurations are equivalent if they represent the same technology up to a common rescaling of all inputs and outputs. The set of equivalence classes $\mathcal{M}_\theta / \sim$ may be regarded as a quotient manifold in which absolute scale has been factored out.

In practice, we approximate this quotient mapping by normalising inputs and outputs and by relying on the latent technology variable $z_i$ learned by $\themodel$, which is encouraged to capture structural rather than purely scale-related variation. Benchmarking in the quotient space amounts to comparing units based on their position in latent technology space rather than on absolute magnitudes of inputs and outputs. In our experiments, we illustrate this idea using Penn World Table macroeconomic data, showing that a standard DEA-based efficiency index exhibits substantial correlation with country size, whereas an index based on $\theframe$ in the quotient space displays a much weaker association.

\subsection{Geometric certification of efficiency scores}
\label{subsec:certification}

Efficiency scores computed from any model can be sensitive to small perturbations in the underlying data or to local irregularities of the estimated frontier. To provide a simple indication of local robustness, we define a certification radius based on the behaviour of the decoder near a given input.

Assume that for fixed latent technology $z$ the decoder mapping $x \mapsto g_\theta(x,z)$ is globally $L_\theta$-Lipschitz in $x$ with respect to the Euclidean norm. Let $J(x)$ denote the Jacobian matrix of partial derivatives of $g_\theta$ with respect to $x$ at $(x,z)$, and let $\sigma_{\min}(J(x))$ be its smallest singular value.

\begin{definition}[Certification radius]
For a given input $x_i$ and latent technology $z_i$, the certification radius is
\[
R_{\mathrm{cert}}(x_i) = \frac{\sigma_{\min}(J(x_i))}{L_\theta}.
\]
\end{definition}

A large value of $R_{\mathrm{cert}}(x_i)$ suggests that the local mapping from inputs to frontier outputs is smooth and relatively well-conditioned, whereas a small value indicates that the decoder may have sharp bends or folds near $x_i$. To interpret this quantity, consider a perturbation $\delta x$ with $\lVert \delta x \rVert_2 \le r < R_{\mathrm{cert}}(x_i)$. By the Lipschitz property, the change in the frontier output is bounded by
\[
\lVert g_\theta(x_i + \delta x, z_i) - g_\theta(x_i, z_i) \rVert_2 \le L_\theta r < \sigma_{\min}(J(x_i)).
\]
Thus, within a ball of radius $r$ in input space, the variation in frontier output is controlled by a bound that is strictly smaller than the smallest local amplification factor implied by the Jacobian. As $\sigma_{\min}(J(x_i))$ decreases towards zero, the certification radius shrinks, signalling that very small changes in inputs may induce large changes in outputs due to local geometric irregularities.

In our empirical analysis, we use $R_{\mathrm{cert}}(x_i)$ as a local robustness indicator for the efficiency score of unit $i$. High nominal efficiency that coincides with a very small certification radius can be interpreted as a ``fragile'' score in the sense that it relies on frontier geometry that is locally ill-conditioned. Such cases may warrant closer scrutiny in regulatory or policy applications. Practical details of Jacobian computation and input whitening are given in Appendix~B.

\section{Experiments}
\label{sec:experiments}

We evaluate $\theframe$ on synthetic data and several real-world domains. The synthetic experiments probe whether $\theframe$ behaves sensibly in classical settings and where it brings structural advantages relative to established frontier estimators. The real-world studies focus on two domains of particular interest for the machine learning and regulatory communities: wind farm operations (WF) and national rail operators (ORR). Additional analyses on urban rail systems (COMET) and macroeconomic data (PWT) are reported in the appendix~C.

Unless otherwise stated, all $\themodel$ models use the same encoder--decoder architecture with domain-specific input/output dimensions and modest hyperparameter tuning. Baselines include DEA with variable returns to scale (VRS), parametric SFA with a Translog specification, a free disposal hull (FDH) estimator, convex nonparametric least squares (CNLS) and a purely predictive machine learning baseline (random forest). Implementation details and hyperparameters are given in Appendix~C.

\subsection{Synthetic experiments}
\label{subsec:synth}

The synthetic experiments examine three stylised settings in which key assumptions commonly imposed in efficiency analysis are selectively violated:

\begin{itemize}
    \item \textbf{Scenario A (non-convex frontier).} A smooth but globally non-convex frontier with saturation effects, designed so that parametric and convex-hull estimators are well-specified or nearly so.
    \item \textbf{Scenario B (heterogeneous technologies).} A mixture of two distinct production technologies, reflecting unobserved technological heterogeneity under a single global input--output space.
    \item \textbf{Scenario C (scale confounding).} A size variable correlated with both inputs and outputs, inducing systematic correlation between estimated efficiency and size for methods that operate purely in observed space.
\end{itemize}

In all scenarios, outputs are generated from known production frontiers with multiplicative inefficiency and noise, following the structural form of the $\themodel$ model. We use $n=500$ DMUs and average results over $30$ Monte Carlo replications.

We compare methods using four metrics: frontier approximation error (Scenario A), inefficiency ranking quality (Scenarios A--C), cluster recovery via adjusted Rand index (Scenario B) and scale bias measured as the correlation between efficiency and size (Scenario C). Full data-generating processes and metric definitions are in Appendix~B.

Table~\ref{tab:synth_summary} summarises the main results. When the data-generating process closely aligns with smooth parametric or low-dimensional predictive models (Scenario A), classical SFA and the machine learning predictor achieve the lowest frontier approximation errors and relatively high inefficiency rank correlations, and $\theframe$ performs comparably. In Scenarios B and C, where technological homogeneity and scale separability are violated, $\theframe$ achieves better recovery of latent technology groups (higher ARI) and substantially attenuates the correlation between estimated inefficiency and size when benchmarking is performed in the quotient space. This suggests that the advantages of $\theframe$ are structural rather than universal, and are most pronounced when heterogeneity and scale confounding are present.

\subsection{Robustness of efficiency scores in national rail (ORR)}
\label{subsec:orr}

We next examine the robustness of frontier-based performance assessments for British train operating companies (TOCs), using panel data compiled by the Office of Rail and Road. The aim is not to establish numerical dominance over classical frontier estimators, but to illustrate how $\theframe$ augments point assessments with robustness diagnostics that are directly relevant for regulatory benchmarking.

\paragraph{Data and setup.}
The ORR data cover multiple TOCs observed over several years. Inputs include labour, route length, rolling stock and planned capacity; outputs include passenger-kilometres and train-kilometres, with additional quality indicators used in supplementary analyses. Route length also serves as a scale proxy. We treat each operator--year as one observation, apply log-transformations for numerical stability and estimate a $\themodel$ model with a low-dimensional technology space. For comparison, we estimate VRS DEA and parametric SFA models using the same inputs and outputs. Further details on variable construction and preprocessing are given in Appendix~E.

\begin{table}[ht]
\centering
\caption{ORR: certification radius percentiles across operator--year observations.}
\label{tab:orr_cert_quantiles}
\begin{center}
    \begin{small}
      \begin{sc}
\begin{tabular}{lccccc}
\hline
Percentile & p0 & p25 & p50 & p75 & p95 \\
\hline
$R_{\mathrm{cert}}$ & 0.105 & 0.227 & 0.293 & 0.340 & 0.392 \\
\hline
\end{tabular}
      \end{sc}
    \end{small}
  \end{center}
  \vskip -0.1in
\end{table}

\begin{figure}[ht]
\vskip 0.2in
  \begin{center}
  \centerline{\includegraphics[width=0.46\columnwidth]{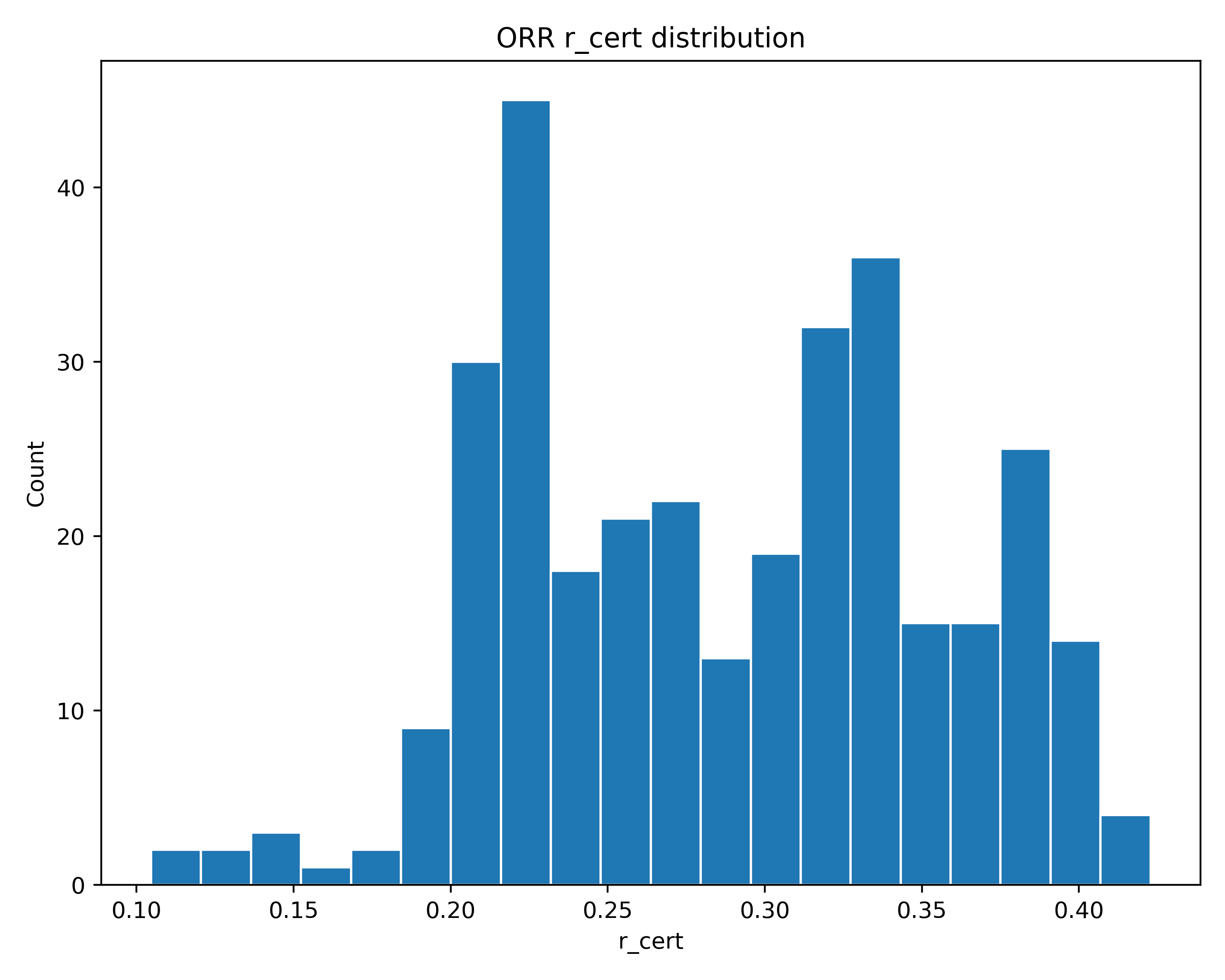}\includegraphics[width=0.48\columnwidth]{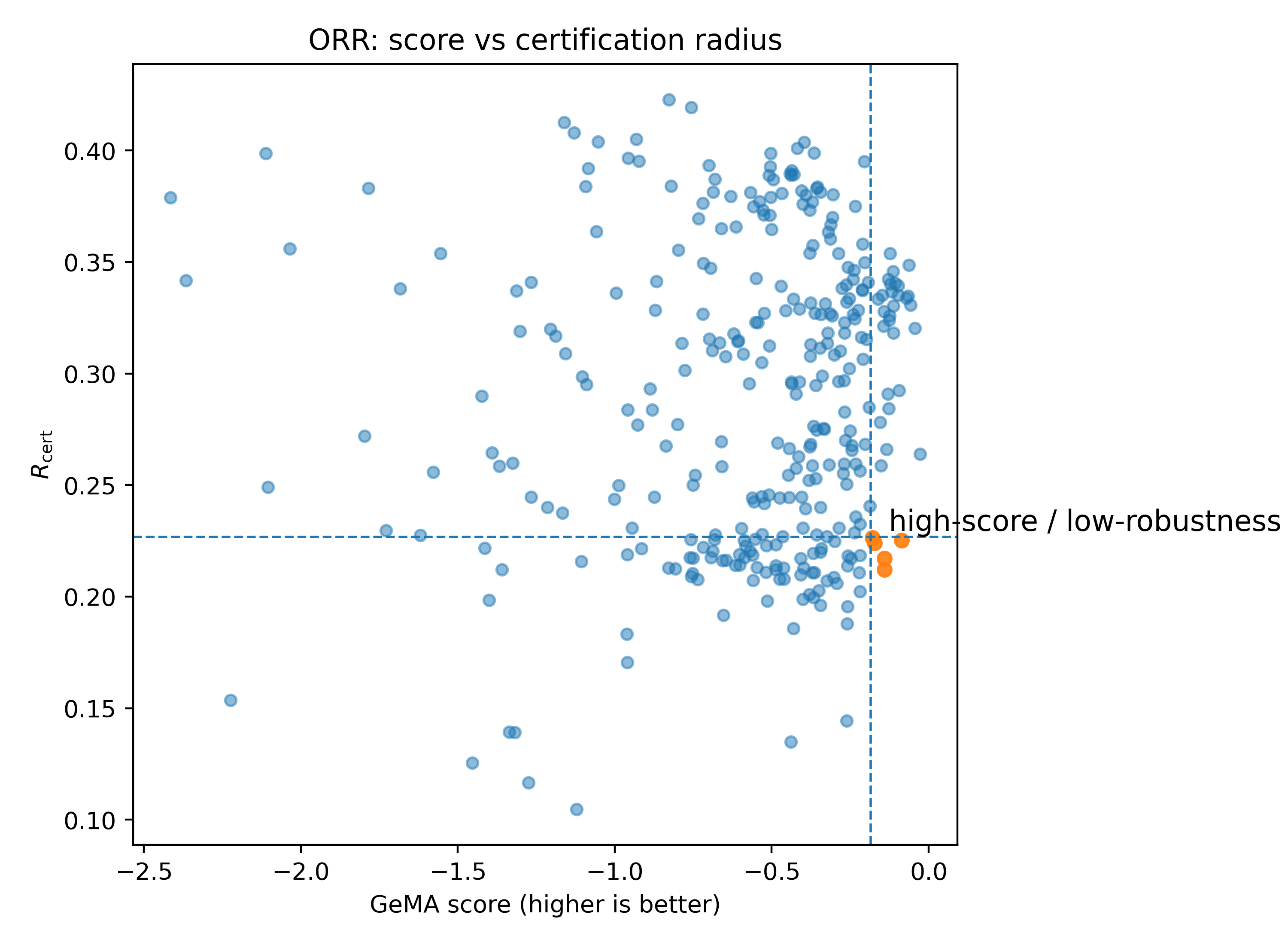}}
    \caption{(Left) Distribution of certification radii $R_{\mathrm{cert}}(\mathbf{x}_i)$ across operator--year observations. A visible left tail indicates cases where performance assessments rely on locally ill-conditioned frontier geometry. (Right)  Model-based score versus certification radius. The dashed lines indicate the top decile of the score and the bottom quartile of $R_{\mathrm{cert}}$, highlighting observations with high point performance but weak robustness guarantees.}
    \label{fig:orr_cert}
  \end{center}
\end{figure}

\paragraph{Certification radii and fragile high scores.}
For each operator--year observation, we compute a certification radius $R_{\mathrm{cert}}(x_i)$ as defined in Section~\ref{subsec:certification}. Figure~\ref{fig:orr_cert} (left) shows the distribution of certification radii; most observations have moderate radii, suggesting locally well-conditioned frontier geometry for a large share of the sample, but the distribution exhibits a non-negligible left tail. Table~\ref{tab:orr_cert_quantiles} reports representative percentiles.

A key use of this diagnostic is to highlight cases where high point scores may be sensitive to noise. Figure~\ref{fig:orr_cert} (right) plots the $\theframe$-based performance score against $R_{\mathrm{cert}}(x_i)$ and marks a ``high-score / low-robustness'' region (top decile of the score combined with bottom quartile of $R_{\mathrm{cert}}$). Observations in this region are not necessarily misclassified, but their rankings are more likely to be fragile with respect to measurement error or small input perturbations. From a regulatory perspective, such cases warrant closer scrutiny than similarly high-scoring observations supported by strong robustness guarantees.

\paragraph{Comparison with classical frontiers.}
Overall rankings under $\theframe$ and SFA/DEA are moderately aligned at the aggregate level, but several TOCs exhibit substantial differences. Some operators that appear highly efficient under SFA have very small certification radii under $\theframe$, indicating scores that rely on locally ill-conditioned frontier geometry, while others have modest efficiency scores but large radii. The certification radius thus adds information that is absent from classical frontier estimators and can help regulators distinguish high scores that are well supported by the frontier geometry from those that are potentially fragile.

\subsection{Non-linear physical frontiers in wind farms (WF)}
\label{subsec:wind}

Our final main case concerns wind farm operations in China and highlights $\theframe$'s ability to recover non-linear physical frontiers in a physics-informed machine learning (PIML) spirit.

\paragraph{Data and model.}
We use a publicly available wind power dataset from the State Grid Renewable Energy Generation Forecasting Competition, covering six wind farms with different turbine models, hub heights and rotor diameters. For each farm, we use two years of 15-minute SCADA measurements, including hub-height wind speed, wind direction, air temperature, air pressure, relative humidity and total active power output; wind direction is encoded via its sine and cosine. We also incorporate farm-level turbine configuration features derived from manufacturer specifications, such as swept rotor area, average hub height, average rotor diameter and number of turbines. All continuous variables are log-transformed or standardised as appropriate; details are in Appendix~E.

We adapt $\theframe$ to the wind domain by treating each 15-minute timestamp as a DMU in a static frontier setting. The model takes as inputs the environmental variables and turbine configuration features and outputs a latent technology representation and a frontier prediction in log-transformed power space. Observations are modelled as $y_{\log} = y_{\log}^{\mathrm{frontier}} - u$, where $u \ge 0$ captures multiplicative operational losses and unexplained deviations from the physical frontier, including wake interactions, curtailment and grid-level constraints, rather than managerial inefficiency in the usual economic sense.

\paragraph{Predictive accuracy and efficiency levels.}
Under a year-based train/validation/test split, $\theframe$ achieves an RMSE of about $0.72$ and an $R^2$ of about $0.82$ on the 2020 test set in log-transformed power space, indicating that the learned frontier is consistent with observed data. Averaging the efficiency ratio $\rho = p_{\mathrm{obs}} / p_{\mathrm{frontier}}$ over time, we find that the six farms operate at roughly $40$--$50\%$ of their learned frontier output once local wind conditions are controlled for, indicating broadly similar operational utilisation across farms.

\begin{figure}[ht]
\vskip 0.2in
  \begin{center}
  \centerline{\includegraphics[width=0.46\columnwidth]{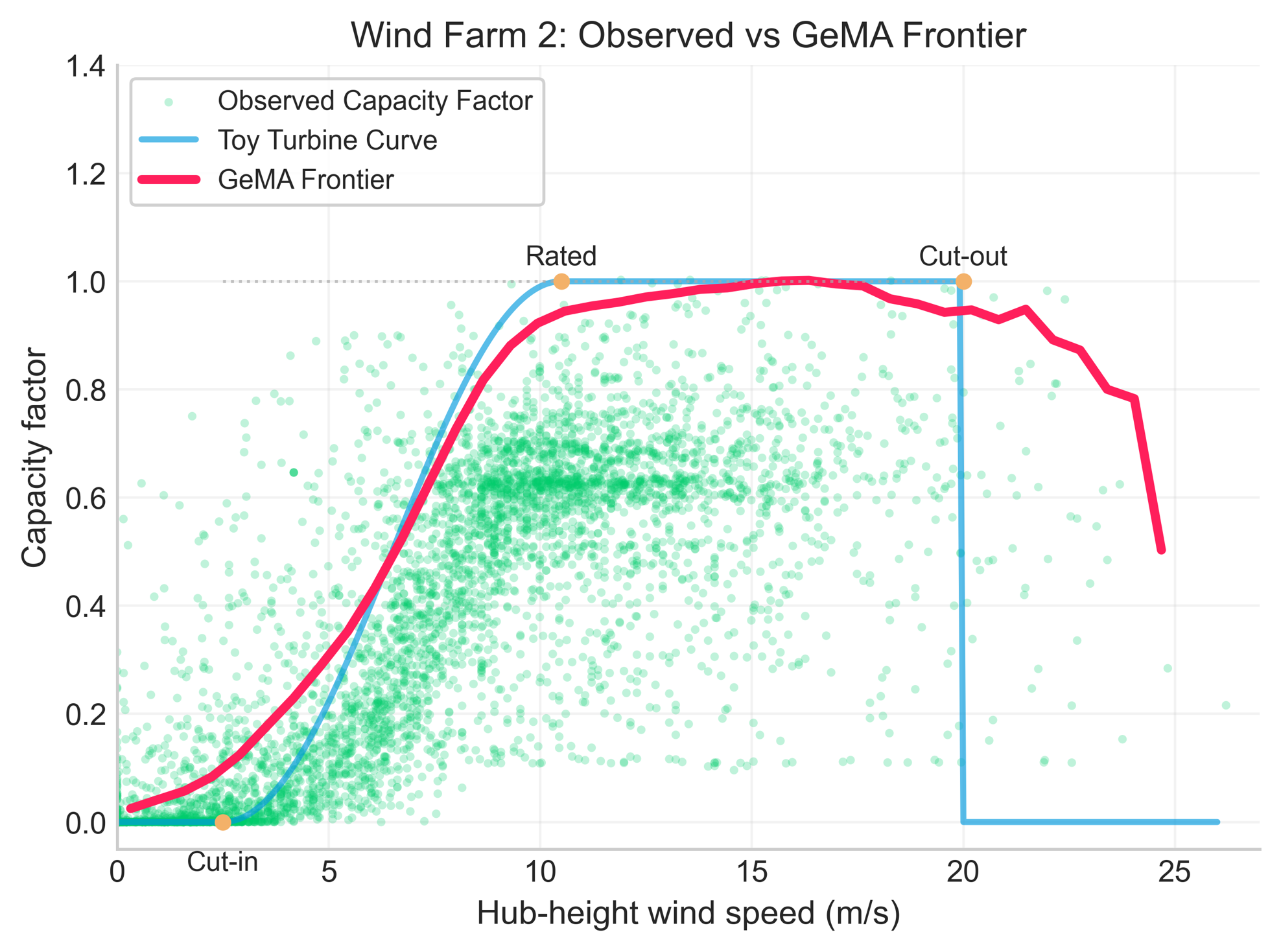}\includegraphics[width=0.48\columnwidth]{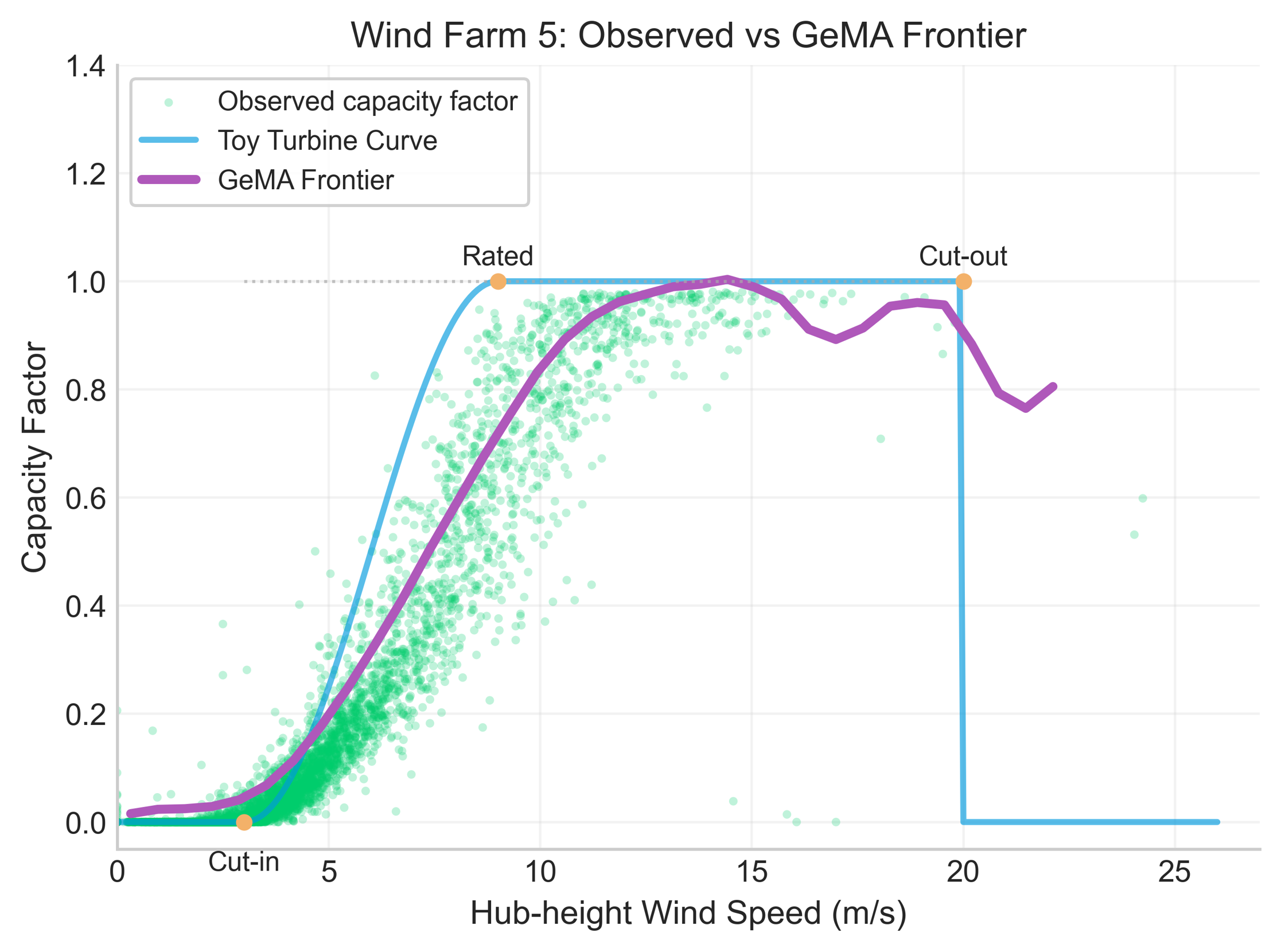}}
    \caption{Wind farms: learned frontiers vs specification-based toy curves for two representative farms (Farm 2 and 5). Scatter points show observed capacity factor versus hub-height wind speed; red curves show the $\theframe$ frontier (normalised by its 95th percentile per farm); blue curves show simple \textit{toy} turbine curves constructed from manufacturer cut-in, rated and cut-out speeds. Orange markers indicate specification-based operating points. (Left) Frontier learned without using the turbine threshold (cut-in/rated speeds) as inputs. (Right) Frontier learned with the turbine threshold (cut-in/rated speeds) included as additional inputs, yielding a tighter alignment with the theoretical plateau and a more pronounced decline near cut-out.}
    \label{fig:wf_powercurves}
  \end{center}
\end{figure}

\paragraph{Learned power curves and comparison to engineering models.}

Figure~\ref{fig:wf_powercurves} shows, for two representative wind farms, the empirical scatter of observed capacity factor versus hub-height wind speed, overlaid with the learned $\theframe$ frontier and a simple ``toy'' turbine curve constructed from manufacturer cut-in, rated and cut-out speeds. To facilitate visual comparison, we normalise the frontier capacity factor for each farm by its $95$th percentile so that the empirical plateau aligns approximately with capacity factor one.

Across all farms, the learned frontiers reproduce the characteristic three-stage structure of turbine power curves: near-zero output below cut-in, a steep non-linear ramp-up in the mid-range and a plateau around rated capacity. At high wind speeds the frontier capacity factor declines instead of remaining flat, particularly in farms with frequent high-wind events, consistent with early curtailment and grid-level constraints that are absent from the idealised toy curves. A simple parametric turbine model fitted to the learned $\theframe$ frontier yields cut-in and rated speeds that cluster around manufacturer values, despite the absence of hard physics constraints in the architecture or loss.

These results indicate that $\theframe$ can recover physically plausible non-linear frontiers from high-frequency operational data, while simultaneously providing an inefficiency factor that captures time-varying operational losses. This illustrates how latent manifold frontiers can be used as flexible approximations to engineering curves in a PIML framework, with potential applications in performance benchmarking, anomaly detection and planning under uncertainty.

\subsection{Additional case studies: COMET and PWT}
\label{subsec:comet_pwt}

For completeness, we briefly summarise two further applications; full details are given in Appendix~G.

\paragraph{Urban rail systems (COMET).}
We apply $\theframe$ to anonymised metro systems from the Community of Metros, covering networks from Asia-Pacific, Europe and the Americas. A two-dimensional latent technology space reveals four endogenous peer groups corresponding to large legacy systems, newer high-density networks and medium-sized balanced systems. Compared with a single global DEA frontier, $\theframe$ provides more graded within-group performance signals and separates structural heterogeneity (peer grouping) from performance differences.

\begin{figure}[ht]
    \centering
    \includegraphics[width=1\columnwidth]{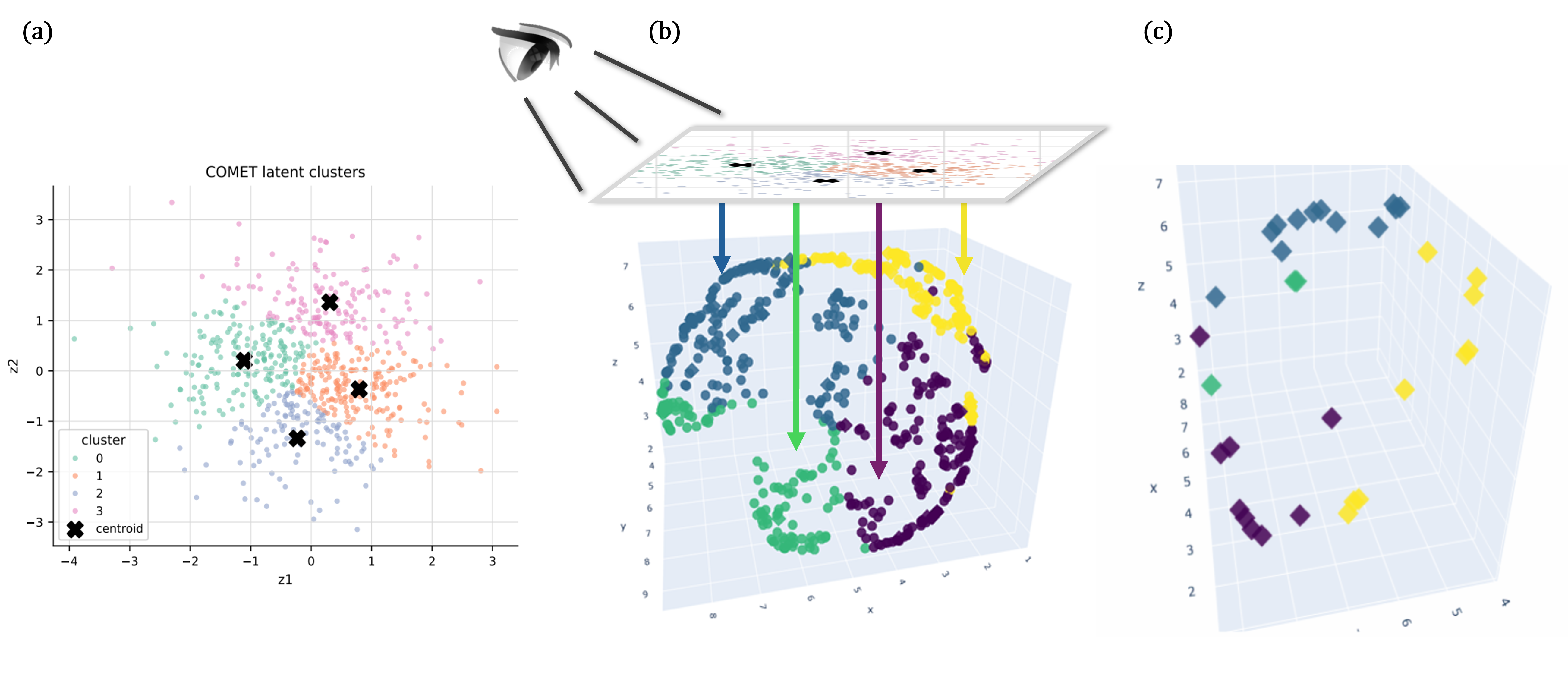}
    \caption{COMET: UMAP embeddings of latent technology vectors $\mathbf{z}_i$ learned by $\theframe$, coloured by GMM cluster assignment ($k=4$). 
    (a) Two-dimensional UMAP projection of the latent technology space used for visualising peer groups. 
    (b) Three-dimensional UMAP view of the same latent manifold, illustrating its overall geometry; the projected points in (a) correspond to this surface. 
    (c) DMUs whose outputs lie on the estimated frontier, shown as the subset of points mapped to the boundary of the learned manifold in latent space. These boundary points represent units operating at the highest efficiency given their latent technology, i.e.\ on the manifold frontier.}
    \label{fig:comet_latent}
\end{figure}

\paragraph{Macroeconomic benchmarking (PWT).}
Using Penn World Table data, we examine how benchmarking in latent technology space alters the interpretation of macroeconomic efficiency. A quotient-based efficiency index defined in the latent manifold attenuates the correlation between efficiency and population size relative to a standard DEA index, and reorders rankings within latent peer groups. This suggests that the quotient construction can mitigate scale-related biases while retaining a familiar frontier-based notion of efficiency.

\section{Discussion and Limitations}
\label{sec:discussion}

Our experiments suggest that $\theframe$ is most informative when production technologies are heterogeneous, non-convex or confounded with scale, and when the goal is to obtain interpretable and robust benchmarking rather than to maximise predictive accuracy alone. In smooth, low-dimensional settings closely aligned with parametric specifications, classical SFA and convex envelopment methods perform very well and $\theframe$ behaves comparably but does not dominate.

Several limitations deserve emphasis. First, model complexity and computational cost are higher than for classical frontier models, as training a deep generative model with monotonicity regularisation and Jacobian-based certification requires GPUs and hyperparameter tuning. Second, while the latent technology and inefficiency variables have clear conceptual roles, the decomposition is not uniquely identifiable in a purely data-driven sense; in practice it is regularised by priors, shape constraints and the SFA-style structural equation, and individual latent dimensions should not be over-interpreted. Third, the quotient construction targets a specific notion of scale equivalence and the certification radius is a qualitative robustness indicator based on conservative Lipschitz bounds rather than a formal adversarial guarantee. These diagnostics should therefore be interpreted in conjunction with domain knowledge and sensitivity checks.

\section{Conclusion}
\label{sec:conclusion}

We introduced Geometric Manifold Analysis ($\theframe$), a latent manifold frontier framework that combines deep generative modelling with classical concepts from efficiency analysis. By modelling the production set as the image of a low-dimensional manifold in joint input--output space and augmenting it with a quotient construction and a certification radius, $\theframe$ provides tools for representing heterogeneous, non-convex technologies and for assessing the local robustness of efficiency scores. Synthetic experiments and case studies on national rail operators, wind farms and macroeconomic data indicate that $\theframe$ behaves sensibly relative to established frontier estimators and can offer additional insight in settings with pronounced heterogeneity, non-convexity or scale bias. Future work includes developing theoretical guarantees for latent manifold frontiers and extending the framework to dynamic ``world-model'' representations of evolving production systems.

\section*{Impact Statement}

This work proposes a latent-manifold framework for efficiency analysis that aims to provide fairer and more interpretable benchmarking of complex systems. In domains such as rail regulation and macroeconomic policy, the ability to distinguish structural heterogeneity from inefficiency and to flag fragile high-efficiency scores may help regulators and policymakers avoid misleading performance assessments. At the same time, the use of deep generative models introduces additional complexity and potential opacity; the latent variables should not be interpreted as causal without further analysis, and the robustness diagnostics we propose are qualitative rather than formal guarantees. We view $\theframe$ as a complementary tool that can support, but not replace, existing domain expertise and established frontier methods.


\bibliography{ICML}
\bibliographystyle{icml2026}

\newpage
\appendix
\onecolumn

\section{Appendix Overview} 

Appendix A provides details of the $\themodel$ likelihood, objective function and network architecture. Appendix B describes the synthetic data‑generating processes used in Section 4. Appendix C lists the data sources, variable definitions and preprocessing steps for the empirical case studies. Appendix D reports model hyperparameters and computational details. Appendices E and F provide additional empirical results and domain‑specific specifications for the wind‑farm and COMET applications, respectively.

\section*{Appendix A: $\themodel$ Model and optimisation}
\label{app:proman}

This appendix gives the explicit form of the $\themodel$ objective used to train $\theframe$. We work in log‑output space and decompose the loss into a reconstruction term and two Kullback–Leibler (KL) penalties for the latent technology and inefficiency variables.

\subsection*{A.1 Likelihood, ELBO, and loss}

For each observation, we work with transformed outputs $\tilde{\mathbf{y}}_i$, typically defined as $\tilde{y} = \log(1 + y)$ for scalar outputs or appropriate log-ratios for macroeconomic data. Given latent variables $(\mathbf{z}_i,u_i)$ and inputs $(\tilde{\mathbf{x}}_i,\mathbf{e}_i,\mathbf{t}_i)$, the decoder produces a log-frontier prediction
$$
    \tilde{\mathbf{y}}_i^{\text{frontier}}
    =
    \mathcal{G}_\theta(\tilde{\mathbf{x}}_i,\mathbf{z}_i,\mathbf{e}_i,\mathbf{t}_i).
$$
The log-space structural equation is
$$
    \tilde{\mathbf{y}}_i = \tilde{\mathbf{y}}_i^{\text{frontier}}-u_i+\bm{\varepsilon}_i,
$$
where $\bm{\varepsilon}_i$ is modelled as Gaussian noise. We denote log‑transformed outputs by $\tilde{\mathbf{y}}$; all equations in this appendix operate in log‑space. 

Assuming independent components with variance $\sigma_y^2$, the likelihood takes the form
$$
    p_\theta(\tilde{\mathbf{y}}_i \mid \tilde{\mathbf{x}}_i,\mathbf{z}_i,u_i)
    =
    \mathcal{N}\bigl(
        \tilde{\mathbf{y}}_i \,\big|\,
        \tilde{\mathbf{y}}_i^{\text{frontier}} - u_i\mathbf{1},
        \sigma_y^2 \mathbf{I}
    \bigr).
$$
In practice, we approximate the negative log-likelihood by a per-output Huber or mean squared error loss between reconstructed and observed log outputs,
$$
    \mathcal{L}_{\mathrm{rec},i}
    =
    \sum_{k=1}^{d_y}
    \ell\bigl( \tilde{y}_{ik}^{\text{rec}} - \tilde{y}_{ik} \bigr),
$$
where $\tilde{\mathbf{y}}_i^{\text{rec}} = \tilde{\mathbf{y}}_i^{\text{frontier}} - u_i$ and $\ell$ is the Huber or squared loss. This choice is convenient, numerically stable and compatible with heavy-tailed deviations from the Gaussian likelihood assumption.

\subsection*{A.2  KL term for latent technology $\mathcal{z}$}

The variational posterior for $\mathbf{z}_i$ is a diagonal Gaussian
$$
    q_\phi(\mathbf{z}_i \mid \cdot)
    = \mathcal{N}\bigl(\bm{\mu}_i^{(z)}, \operatorname{diag}(\bm{\sigma}_i^{2(z)})\bigr),
$$
with prior $p(\mathbf{z}_i) = \mathcal{N}(\mathbf{0},\mathbf{I})$. The KL divergence has the standard closed form
$$
    \mathrm{KL}\bigl(q_\phi(\mathbf{z}_i \mid \cdot)\,\|\,p(\mathbf{z}_i)\bigr)
    =
    -\tfrac{1}{2}
    \sum_{j=1}^{K}
    \left(
        1 + \log \sigma_{ij}^{2(z)}
        - \bigl(\mu_{ij}^{(z)}\bigr)^2
        - \sigma_{ij}^{2(z)}
    \right).
$$
This term encourages the approximate posterior for $\mathbf{z}_i$ to remain close to the standard normal prior, preventing the latent technology space from collapsing or overfitting individual observations.

\subsection*{A.3 KL term for inefficiency $u$}
\label{app:kl_u}

The variational posterior for $\log u_i$ is Gaussian with mean $\mu_i^{(u)}$ and log-variance $\log \sigma_i^{2(u)}$,
$$
    q_\phi(\log u_i \mid \cdot)
    =
    \mathcal{N}\bigl(\mu_i^{(u)}, \sigma_i^{2(u)}\bigr),
$$
so that $u_i$ is log-normal under $q_\phi$:
$$
    q_\phi(u_i \mid \cdot)
    =
    \mathrm{LogNormal}\bigl(u_i \mid \mu_i^{(u)}, \sigma_i^{2(u)}\bigr),
    \qquad u_i \ge 0.
$$
The prior for $u_i$ is exponential with rate $\lambda$, $$p(u_i) = \lambda \exp(-\lambda u_i), \quad u_i \ge 0.$$

The KL divergence between the log-normal posterior and exponential prior can be written as
$$
    \mathrm{KL}\bigl(q_\phi(u_i \mid \cdot)\,\|\,p(u_i)\bigr)
    =
    - H\bigl(q_\phi(u_i \mid \cdot)\bigr)
    + \log \lambda
    + \lambda\,\mathbb{E}_{q_\phi}[u_i],
$$
where $H(q)$ is the differential entropy of the log-normal distribution and $\mathbb{E}_{q_\phi}[u_i]$ its mean. For $u \sim \mathrm{LogNormal}(\mu,\sigma^2)$ these quantities are
$$
    H(q) = \mu + \tfrac{1}{2}\log(2\pi e \sigma^2),
    \qquad
    \mathbb{E}[u] = \exp\bigl(\mu + \tfrac{1}{2}\sigma^2\bigr).
$$
Substituting into the expression above yields

$$
    \mathrm{KL}\bigl(q_\phi(u_i \mid \cdot)\,\|\,p(u_i)\bigr)
    =
    -\mu_i^{(u)} - \tfrac{1}{2}\log\bigl(2\pi e \sigma_i^{2(u)}\bigr)
    + \log \lambda
    + \lambda \exp\bigl(\mu_i^{(u)} + \tfrac{1}{2}\sigma_i^{2(u)}\bigr).
$$

In implementation, for numerical stability, we parameterise $\sigma_i^{2(u)}$ via its log-variance and the rate $\lambda$ via a learnable $\log \lambda$ passed through a softplus function to ensure positivity.

\subsection*{A.4 Overall training objective}

Aggregating over observations, the $\themodel$ objective can be written as

$$
    \mathcal{L}(\theta,\phi)
    =
    \sum_{i=1}^{N}
    \left\{
        \mathcal{L}_{\mathrm{rec},i}
        + \beta\,\mathrm{KL}\bigl(q_\phi(\mathbf{z}_i \mid \cdot)\,\|\,p(\mathbf{z}_i)\bigr)
        + \gamma\,\mathrm{KL}\bigl(q_\phi(u_i \mid \cdot)\,\|\,p(u_i)\bigr)
    \right\}
    +
    \lambda_{\mathrm{mono}}\,\mathcal{R}_{\mathrm{mono}}(\theta),
$$

where $\beta$ and $\gamma$ weight the KL terms and $\lambda_{\mathrm{mono}}$ controls the strength of monotonicity regularisation. This objective corresponds to Equation 4 in the main text, with $\beta$ and $\gamma$ controlling the relative strength of the KL terms and  $\lambda_{\mathrm{mono}}$ the monotonicity regulariser introduced in Section 3.3.2.

We employ a simple annealing schedule for $\beta$, increasing it from zero to one over a fixed number of epochs. The objective is minimised using Adam with early stopping based on validation error and a modest dropout schedule in early epochs.

\section*{Appendix A (continued): Network architecture and regularisation}
\label{app:model}

This appendix provides additional details of the $\themodel$ encoder and decoder parameterisation, the training objective and the monotonicity regulariser used across all domains.

\subsection*{A.5 Encoder and decoder parameterisation}

The encoder implements the approximate posterior $q_\phi(\mathbf{z}_i,u_i \mid \mathbf{x}_i,\mathbf{y}_i)$ in Section 3.3.1. And the encoder network receives concatenated inputs and outputs $(\mathbf{x}_i,\mathbf{y}_i)$ and produces parameters of approximate posterior distributions for the latent variables $(\mathbf{z}_i,u_i)$. Concretely, we use a feedforward neural network with several hidden layers and two heads:
\begin{itemize}
    \item a \emph{technology head} that outputs a mean vector $\bm{\mu}_i^{(z)} \in \mathbb{R}^K$ and a log-variance vector $\log \bm{\sigma}_i^{2(z)} \in \mathbb{R}^K$, defining a diagonal Gaussian
    $
        q_\phi(\mathbf{z}_i \mid \mathbf{x}_i,\mathbf{y}_i)
        =
        \mathcal{N}\bigl(\bm{\mu}_i^{(z)}, \operatorname{diag}(\bm{\sigma}_i^{2(z)})\bigr);
    $
    \item an \emph{inefficiency head} that outputs parameters for $\log u_i$, which is taken to be Gaussian and then mapped to $u_i \ge 0$ via an exponential or softplus transformation.
\end{itemize}

Sampling from $q_\phi(\mathbf{z}_i,u_i \mid \mathbf{x}_i,\mathbf{y}_i)$ is implemented using the reparameterisation trick, with independent standard normal noise transformed by the encoder outputs. This allows gradients to propagate through stochastic nodes during training.

The decoder network $\mathcal{G}_\theta$ receives the concatenated inputs $\mathbf{x}_i$ and latent technology $\mathbf{z}_i$ and outputs a frontier prediction $\mathbf{y}_i^\ast = \mathcal{G}_\theta(\mathbf{x}_i,\mathbf{z}_i)$. It is parameterised as a feedforward network with several hidden layers and an output layer of dimension $d_y$. Activation functions are chosen to ensure non-negativity of outputs where required by the application domain.

The hidden layers use smooth non-linear activations GELU (Gaussian Error Linear Unit) for COMET and ORR, SiLU (Sigmoid-weighted Linear Unit) for PWT and WF, with a softplus or linear activation in the output layer depending on the range of the target variables.

\subsection*{A.6 Monotonicity regularisation}

The monotonicity regularisation term penalises violations of weak monotonicity in inputs. For a set of reference points in input space $\{\mathbf{x}^{(g)}\}$ and unit vectors $\mathbf{e}_j$ in the input directions, we approximate partial derivatives of the decoder outputs with respect to inputs via finite differences. For a small step $\delta > 0$, we compute
$
    \Delta_{j}^{(g)}(\theta)
    =
    \mathcal{G}_\theta(\mathbf{x}^{(g)} + \delta \mathbf{e}_j,\mathbf{z})
    - \mathcal{G}_\theta(\mathbf{x}^{(g)},\mathbf{z}),
$
for representative values of $\mathbf{z}$. The regularisation term is
$
    \mathcal{R}_{\mathrm{mono}}(\theta)
    =
    \sum_{g,j}
    \bigl\| \min(0, \Delta_{j}^{(g)}(\theta)) \bigr\|_1,
$
so that negative increments in the output when an input increases are penalised. In practice, this term is estimated stochastically using minibatches of data and latent samples, which keeps the additional computational cost modest. The resulting penalty $\mathcal{R}_{\mathrm{mono}}(\theta)$ is scaled by $\lambda_{\mathrm{mono}}$ in the training objective and is estimated using minibatches and randomly sampled finite‑difference directions.

\subsection*{A.7 Entity and time embeddings}

For panel data with entity and time identifiers, we map each unique entity and each unique time period to a low-dimensional embedding vector. These embeddings are concatenated with transformed inputs and outputs at the encoder, and with transformed inputs at the decoder. These embeddings allow the model to capture persistent cross‑sectional and temporal patterns without explicitly specifying a state‑space evolution model. This allows $\themodel$ to capture persistent cross-sectional differences and coarse temporal effects without explicitly modelling state evolution. The same architecture is used for COMET, ORR, PWT and the wind farm dataset, with only the embedding sizes and input dimensions varying across domains.

\subsection*{A.8 Input whitening and Jacobian computation}

For the computation of certification radii in Section~\ref{subsec:certification}, we stabilise the Jacobian calculation by working with whitened inputs. Let $\bm{\mu}_X$ and $\Sigma_X$ denote the empirical mean vector and covariance matrix of the transformed inputs over the training set. We form a whitening matrix $W$ from the Cholesky factor of $\Sigma_X + \epsilon I$, with a small diagonal regulariser $\epsilon > 0$, and define
$$
    \tilde{\mathbf{x}}^{\mathrm{white}}
    =
    W (\tilde{\mathbf{x}} - \bm{\mu}_X).
$$

The decoder $\mathcal{G}_\theta$ is composed with this whitening transformation when computing Jacobians: we differentiate $\mathbf{x} \mapsto \mathcal{G}_\theta(W(\mathbf{x}-\bm{\mu}_X),\mathbf{z})$ with respect to $\mathbf{x}$ using automatic differentiation. Singular values of the resulting Jacobian $J(\mathbf{x})$ are obtained via standard linear algebra routines. Combined with spectral normalisation of decoder layers, this yields conservative but numerically stable estimates of $\sigma_{\min}(J(\mathbf{x}))$ and hence of the certification radius $R_{\mathrm{cert}}(\mathbf{x})$. These spectral-norm-based estimates are conservative and are intended as qualitative indicators of local conditioning rather than formal worst‑case guarantees

\section*{Appendix B: Synthetic data-generating processes}
\label{app:dgp}

This appendix describes the data-generating processes (DGPs) used in the synthetic experiments of Section~\ref{subsec:synth}. We keep the notation consistent with the main text:$\mathbf{x}_i$ denotes inputs, $\mathbf{y}_i$ outputs, $\mathbf{u}_i$ inefficiency and $f$ the true frontier. Baseline methods and evaluation metrics are defined in the main text and are not repeated here.

\subsection*{B.1 Monte Carlo design}

For each scenario, we generate $n=500$ decision-making units (DMUs) and repeat the experiment over 30 Monte Carlo replications, using different random seeds. Reported results in the main text correspond to averages across these replications, with standard deviations reported in parentheses. All performance metrics in Table 1 are computed by averaging over 30 replications of each scenario, with standard deviations in parentheses.

\subsection*{B.2 Scenario A: Non-convex frontier}

Scenario~A generates a smooth but globally non-convex frontier over two inputs. We draw
$\mathbf{x}_i=(x_{i,1},x_{i,2})$ i.i.d.\ from $\mathrm{Unif}[0,1]^2$.
The true frontier is
\[
f^\ast(\mathbf{x}_i)
=
a\bigl(1-e^{-b x_{i,1}}\bigr)\bigl(1-e^{-b x_{i,2}}\bigr)
+
0.2\exp\!\left(
-\frac{(x_{i,1}-0.5)^2+(x_{i,2}-0.2)^2}{0.02}
\right),
\]
where the first term produces saturation effects while the second term introduces a local non-convex ``bump''.
Observed output follows the multiplicative structure
\[
y_i
=
f^\ast(\mathbf{x}_i)\exp(-u_i)\exp(\varepsilon_i),
\qquad
u_i \sim \mathrm{HalfNormal}(0,\sigma_u^2),\ 
\varepsilon_i \sim \mathcal{N}(0,\sigma_\varepsilon^2),
\]
with $\sigma_u=0.3$ and $\sigma_\varepsilon=0.05$ in the baseline calibration (and $a=1$, $b=2$).

This specification yields a smooth but globally non‑convex frontier with a local ‘bump’ that is difficult to approximate using globally convex or low‑order parametric forms.

\subsection*{B.3 Scenario B: Heterogeneous technologies}

Scenario~B introduces unobserved technological heterogeneity through a mixture of two distinct production functions.
We draw $\mathbf{x}_i$ i.i.d.\ from $\mathrm{Unif}[0.1,2.0]^2$ and assign group labels
$g_i\in\{1,2\}$ independently with equal probability.
Conditional on $g_i$, the true frontier is
\[
f^\ast(\mathbf{x}_i)=
\begin{cases}
A x_{i,1}^{\alpha_1}x_{i,2}^{\alpha_2}, & g_i=1,\\[4pt]
B\left(\delta x_{i,1}^{\rho}+(1-\delta)x_{i,2}^{\rho}\right)^{1/\rho}, & g_i=2,
\end{cases}
\]
where the first component is Cobb--Douglas and the second is a CES technology.
In the baseline calibration, $(A,\alpha_1,\alpha_2)=(1,0.4,0.6)$ and $(B,\delta,\rho)=(1.1,0.3,-0.5)$.
Outputs are generated as
\[
y_i
=
f^\ast(\mathbf{x}_i)\exp(-u_i)\exp(\varepsilon_i),
\qquad
u_i \sim \mathrm{HalfNormal}(0,\sigma_u^2),\ 
\varepsilon_i \sim \mathcal{N}(0,\sigma_\varepsilon^2),
\]
with $\sigma_u=0.25$ and $\sigma_\varepsilon=0.05$.

This mixture design reflects unobserved technological regimes governed by distinct production functions.

\subsection*{B.4 Scenario C: Scale confounding}

Scenario~C generates a scale variable that is correlated with both inputs and outputs.
We draw $\log s_i \sim \mathcal{N}(0,1)$ and set $s_i=\exp(\log s_i)$.
We then generate baseline inputs $\tilde{\mathbf{x}}_i \sim \mathrm{Unif}[0.5,1.5]^2$ and define observed inputs
\[
\mathbf{x}_i = s_i\,\tilde{\mathbf{x}}_i.
\]
The true frontier allows size to affect output both through scaled inputs and through an additional multiplicative term:
\[
f^\ast(\mathbf{x}_i,s_i)
=
\theta\, s_i^{\gamma}\, x_{i,1}^{\alpha_1}x_{i,2}^{\alpha_2},
\]
with baseline calibration $\theta=1$, $(\alpha_1,\alpha_2)=(0.3,0.4)$ and $\gamma=0.3$.
Observed output is generated as
\[
y_i
=
f^\ast(\mathbf{x}_i,s_i)\exp(-u_i)\exp(\varepsilon_i),
\qquad
u_i \sim \mathrm{HalfNormal}(0,\sigma_u^2),\ 
\varepsilon_i \sim \mathcal{N}(0,\sigma_\varepsilon^2),
\]
with $\sigma_u=0.3$ and $\sigma_\varepsilon=0.06$.

This design induces a strong correlation between size and both inputs and outputs, leading to systematic scale‑related bias for methods that operate purely in observed input–output space.

\subsection*{B.5 Design interpretation}

\begin{table}[ht]
\centering
\caption{Synthetic experiments: Monte Carlo summary of frontier approximation error and rank correlation with true inefficiency across methods and scenarios. Means with standard deviations in parentheses.}
\label{tab:synth_summary}

\begin{tabular}{lccc}
\hline
Method & Scenario A & Scenario B & Scenario C \\
& Frontier error / Rank corr. & ARI / Rank corr. & Size corr. / Rank corr. \\
\hline
DEA (VRS) & - / 0.655 (0.042) & 0.782 (0.024) & - / 0.801 (0.023) \\
SFA (Translog) & 0.216 (0.009) / 0.727 (0.041) & 0.803 (0.017) & 0.038 (0.022) / 0.818 (0.014) \\
FDH & - / 0.595 (0.039) & 0.701 (0.028) & - / 0.726 (0.027) \\
CNLS & 10.395 (38.537) / 0.350 (0.041) & 0.506 (0.039) & 0.154 (0.064) / 0.786 (0.056) \\
ML predictor (RF) & 0.405 (0.719) / 0.780 (0.023) & 0.767 (0.028) & 0.093 (0.024) / 0.778 (0.023) \\
\theframe & \textbf{0.287 (0.061)} / \textbf{0.804 (0.018)} & \textbf{0.862 (0.019)} & \textbf{0.021 (0.018)} / \textbf{0.832 (0.012)} \\
\hline
\end{tabular}
\end{table}

The three scenarios are designed to isolate distinct assumption violations commonly encountered in efficiency analysis. Scenario~A violates global convexity while maintaining smoothness, providing a setting in which flexible parametric models may perform well. Scenario~B introduces unobserved technological heterogeneity through fundamentally different production functions, challenging single-frontier approaches. Scenario~C induces strong scale confounding by allowing size to affect output both through scaled inputs and through the production frontier itself.

The simulations are not intended to establish universal numerical dominance, but to evaluate whether methods behave sensibly under classical conditions and whether structural advantages emerge when convexity, homogeneity, or scale separability assumptions are violated.

\section*{Appendix C: Data sources, variable definitions and preprocessing}
\label{app:data}

\subsection*{C.1 Data availability and references}

The COMET datasets are proprietary and used under non-disclosure agreements; aggregate statistics and derived quantities are reported in anonymised form. The GB national rail dataset is publicly released by the Office of Rail and Road (ORR) at \url{https://dataportal.orr.gov.uk}. The Penn World Table (PWT) data are publicly available from \citet{feenstra2015next} at \url{www.ggdc.net/pwt}. The wind farm data are based on the Chinese State Grid hosting the Renewable Energy Generation Forecasting Competition and the derived dataset described in \citet{chen2022solar}, available at \url{https://github.com/Bob05757/Renewable-energy-generation-input-feature-variables-analysis} under Creative Commons Attribution 4.0 International License. Code and preprocessing scripts for the public datasets will be released upon publication.

\subsection*{C.2 COMET (urban rail systems) variables and roles}

We use data for the years 1994–2019, treating each (operator, year) as one observation. Table~\ref{tab:vars-comet} lists the variables used in the COMET case study. All continuous variables are transformed by $\log(1+x)$ before standardisation, as described in Appendix C.6.

\begin{table}[ht]
\centering
\caption{COMET variables: definitions and roles.}
\label{tab:vars-comet}

\begin{tabular}{llll}
\hline
Name & Description & Role & Transformation \\
\hline
Operator      & Metro system identifier                 & Entity ID & -- \\
Year          & Calendar year              & Time ID   & -- \\
Staff         & Number of staff (FTE)                   & Input     & $\log(1+x)$ \\
Capacity      & Rolling stock seating capacity (car seats) & Input     & $\log(1+x)$ \\
Stations      & Number of stations                      & Input     & $\log(1+x)$ \\
RouteLength   & Network Length (km)                     & Input/Scale & $\log(1+x)$ \\
Fleets        & Total number of cars   & Input     & $\log(1+x)$ \\
PassKm        & Annual passenger-kilometres             & Output    & $\log(1+x)$ \\
CarKm         & Annual car-kilometres        & Output    & $\log(1+x)$ \\
\hline
\end{tabular}
\end{table}

\subsection*{C.3 ORR (GB rail operators) variables and roles}

We use data for years 2000–2020, treating each (operator, year) as one observation. Table~\ref{tab:vars-orr} summarises the variables used in the ORR case study. All continuous variables are transformed by $\log(1+x)$ before standardisation, as described in Appendix C.6..

\begin{table}[ht]
\centering
\caption{ORR variables: definitions and roles.}
\label{tab:vars-orr}
\begin{tabular}{llll}
\hline
Name & Description & Role & Transformation \\
\hline
Operator           & Train operating company ID          & Entity ID & -- \\
Year               & Calendar year                       & Time ID   & -- \\
Labour     & Number of staff (FTE)    & Input     & $\log(1+x)$ \\
Route     & Route kilometres operated by operators                   & Input/Scale & $\log(1+x)$ \\
Stock      & Train rollingstock in service   & Input     & $\log(1+x)$ \\
Plan       & Numbers of trains planned & Input  & $\log(1+x)$ \\
Station    & Number of stations managed by operator           & Input     & $\log(1+x)$ \\
Pkm       & Passenger kilometres by operator (billion)                & Output    & $\log(1+x)$ \\
PTkm      & Passenger train kilometres by operator         & Output    & $\log(1+x)$ \\
PJ        & Passenger journeys by operator (million)                  & Output (aux) & $\log(1+x)$ \\
PPM       & Public Performance Measure      & Output (aux) & $\log(1+x)$ \\
CaSL     & Cancellations and Significant Lateness & Output (aux) & $\log(1+x)$\\
\hline
\end{tabular}
\end{table}

Public Performance Measure (PPM) is a measure of the percentage of trains arriving on time. A train is defined as on time if it arrives at its final destination within ten minutes of the planned arrival time for long-distance services, and within five minutes for all other services.

\subsection*{C.4 PWT variables and roles}

We use data for the years 1970–2019, treating each (country, year) as one observation. We use Penn World Table data (version 10.0, \citep{feenstra2015next}) with country-year observations. Table~\ref{tab:vars-pwt} lists the main variables. All continuous variables are log-transformed after constructing per-worker or per-hour quantities where applicable, as described in Appendix C.6.

\begin{table}[ht]
\centering
\caption{PWT variables: definitions and roles.}
\label{tab:vars-pwt}
\begin{tabular}{llll}
\hline
Name & Description & Role & Transformation \\
\hline
country & Country code / identifier         & Entity ID & -- \\
year    & Calendar year                     & Time ID   & -- \\
rkna    & Capital services at constant 2017 national prices  & Input     & $\log(\text{rkna}/\text{emp})$ \\
hc      & Human capital index      & Input     & $\log(\text{hc})$ \\
emp     & Number of persons engaged (millions) & Input & $\log(\text{emp})$ \\
rgdpo   & Output-side real GDP at chained PPPs (million) & Output & $\log(\text{rgdpo}/\text{hours})$ \\ 
pop     & Population (million)
                        & Scale     & $\log(\text{pop})$ \\
\hline
\end{tabular}
\end{table}
We normalise capital and output by employment or hours where data permit, following standard growth accounting practice.

\subsection*{C.5 Wind farms (WF) variables and roles}

We use two years of 15-minute SCADA data (2019–2020), aggregating or filtering as described in Section~\ref{app:wind_specs}. Table~\ref{tab:vars-wf} summarises the variables used in the wind farm case study. All continuous variables are transformed by $\log(1+x)$ before standardisation, as described in Appendix C.6.

\begin{table}[ht]
\centering
\caption{WF variables: definitions and roles.}
\label{tab:vars-wf}
\begin{tabular}{llll}
\hline
Name & Description & Role & Transformation \\
\hline
asset\_num           & Wind farm identifier             & Entity ID & -- \\
year                 & Calendar year                    & Time ID   & -- \\
ws\_hub              & Hub-height wind speed           & Input     & $\log(1+x)$ \\
temp\_air            & Air temperature                  & Input     & $\log(1+x)$ \\
pressure\_air        & Air pressure                     & Input     & $\log(1+x)$  \\
rel\_humidity        & Relative humidity                & Input     & $\log(1+x)$ \\
cos\_dir\_hub        & $\cos(\text{wind direction})$    & Input     & standardised \\
sin\_dir\_hub        & $\sin(\text{wind direction})$    & Input     & standardised \\
swept\_area          & Capacity-weighted swept rotor area & Input   & $\log(1+x)$ \\
hub\_height\_avg     & Average hub height               & Input     & $\log(1+x)$ \\
rotor\_diameter\_avg & Average rotor diameter           & Input     & $\log(1+x)$ \\
num\_turbines        & Number of turbines               & Input     & $\log(1+x)$ \\
power\_density\_1\_avg & Nominal power / swept area     & Input     & $\log(1+x)$ \\
power\_density\_2\_avg & Swept area / nominal power     & Input     & $\log(1+x)$ \\
nominal\_capacity    & Installed capacity (MW)          & Scale     & $\log(1+x)$ \\
power                & Total active power output (MW)   & Output    & $\log(1+x)$ \\
\hline
\end{tabular}
\end{table}

\subsection*{C.6 Dataset-specific preprocessing}

Each empirical dataset requires modest preprocessing, but we adhere to a consistent set of principles:

\begin{itemize}
    \item \emph{Transformations.} Inputs and outputs that span several orders of magnitude are log-transformed using $\log(1+x)$ to stabilise variance. Ratio variables (such as per-capita quantities) are formed prior to log transformation where appropriate.
    \item \emph{Normalisation.} After transformation, continuous variables are standardised to zero mean and unit variance based on the training set; the same scaling is applied to validation and test sets.
    \item \emph{Missing values.} Observations with missing key inputs or outputs are removed if they are rare; otherwise, we use simple imputation schemes (such as median imputation) and include an indicator variable where necessary. For the wind farm dataset, short gaps in the time series are left as missing and excluded from training to avoid introducing spurious patterns.
    \item \emph{Panel structure.} For panel datasets such as ORR and PWT, entity and time identifiers are encoded via learned embeddings as described in Appendix~\ref{app:model}, allowing the model to capture persistent and temporal effects without explicit state-space structure.
\end{itemize}

These implementation details are held fixed across the synthetic and empirical studies unless otherwise noted, ensuring that differences in behaviour reflect structural properties of the datasets and model rather than ad hoc tuning.

\subsection*{C.7 UMAP projections and clustering in latent space}

For visualising latent technology spaces and identifying peer groups (Sections~\ref{subsec:comet_pwt}), we use UMAP \citep{mcinnes2018umap, healy2024uniform} to project the latent vectors $\mathbf{z}_i$ into two dimensions. Unless otherwise stated, we adopt the following hyperparameters:

\begin{itemize}
    \item number of neighbours set between $15$ and $50$, depending on dataset size;
    \item minimum distance parameter in $[0.1, 0.3]$ to balance local detail and global structure;
    \item Euclidean distance in latent space as the base metric.
\end{itemize}

After projection, we apply $k$-means clustering with $k$ chosen by a combination of the elbow heuristic and qualitative inspection of cluster stability \citep{sinaga2020unsupervised}. In practice, the latent manifolds for COMET and ORR admit a small number of interpretable clusters (typically between three and five), which we interpret as endogenous peer groups for benchmarking. We do not attempt to optimise clustering hyperparameters exhaustively, as the qualitative structure is robust across reasonable choices.

\section*{Appendix D: Hyperparameters and computational setup}
\label{app:hparams}

\subsection*{D.1 Model hyperparameters across domains}

Table~\ref{tab:hparams-main} summarises the main $\themodel$ hyperparameters used across domains. Values are obtained by coarse hyperparameter search on validation data; results were robust within the indicated ranges.

\begin{table}[ht]
\centering
\caption{$\themodel$ hyperparameters by domain.}
\label{tab:hparams-main}
\begin{tabular}{lcccccc}
\hline
Domain & $K$ (tech dim) & Hidden dim & Epochs & Batch size & Learning rate & $\gamma_u$ \\
\hline
WF     & 2 & 192 & 300 & 1024 & $5\times 10^{-4}$ & 1.0 \\
PWT    & 4 & 192 & 200 & 256  & $3.8\times 10^{-3}$ & 0.213 \\
ORR    & 4 & 128 & 200 & 1792 & $4.7\times 10^{-3}$ & 0.223 \\
COMET  & 2 & 128 & 200 & 2048 & $4.4\times 10^{-3}$ & 0.201 \\
\hline
\end{tabular}

\end{table}

Hyperparameters were selected via coarse search on validation data; we found that performance was stable within the ranges indicated

\begin{table}[ht]
\centering
\caption{Additional architectural and regularisation choices.}
\label{tab:hparams-arch}
\begin{tabular}{lcccc}
\hline
Domain & Activation & Spectral norm & Mono weight & Ranking loss weight \\
\hline
WF     & \texttt{SILU} & No  & 0.0        & (if used; else 0) \\
PWT    & \texttt{SILU} & (default) & $4.8\times 10^{-4}$ & 0.271 \\
ORR    & \texttt{GELU} & Yes & $2.2\times 10^{-4}$ & 0.298 \\
COMET  & \texttt{GELU} & Yes & $1.0\times 10^{-4}$ & 0.056 \\
\hline
\end{tabular}
\end{table}

\subsection*{D.2 Training schedules and regularisation}

All $\themodel$ models are trained using Adam \citep{kingma2014adam} with mini-batches of size between $128$ and $512$ and an initial learning rate in $[10^{-4}, 5\times 10^{-4}]$. We employ early stopping based on validation reconstruction loss, with a patience window of $20$ to $50$ epochs depending on dataset size. The KL weight $\beta$ on the latent technology term is annealed linearly from zero to one over the first $20$ epochs, which encourages the model to find a good reconstruction before fully regularising the latent space.

Dropout and weight decay are used to mitigate overfitting. For the encoder and decoder we apply dropout rates between $0.05$ and $0.2$ in the hidden layers during early training; dropout is reduced or turned off once validation performance stabilises. Weight decay is set in $[10^{-6}, 10^{-4}]$. For the monotonicity regulariser $\mathcal{R}_{\mathrm{mono}}(\theta)$, we set $\lambda_{\mathrm{mono}}$ to a small value (for example $10^{-3}$ to $10^{-2}$) and monitor the fraction of finite-difference violations during training to ensure that the constraint is effective without dominating the objective. These settings are intended to provide robust training across domains rather than dataset‑specific optimisation

\subsection*{D.3 Computational setup}
All experiments were run on a single-node machine with 4 CPU cores, 32~GB RAM and one NVIDIA Quadro RTX~6000 GPU. Mixed-precision training (\texttt{use\_amp=true}) was enabled for the COMET, ORR and PWT experiments. Depending on dataset size and architecture, training a single $\themodel$ instance took between approximately 0.5 and 3 hours wall-clock time.

We use the early-stopping, dropout-annealing and Jacobian-whitening settings described in Appendix A.8, with domain-specific values given in the configuration files.

We acknowledge computational resources and support provided by the Imperial College Research Computing Service (http://doi.org/10.14469/hpc/2232).

\section*{Appendix E: Wind farm specifications and additional results}
\label{app:wind_specs}

This appendix reports technical specifications of the wind farms and turbines used in Section~\ref{sec:experiments}, together with additional parameter-level comparisons between specification-based turbine curves and the learned $\theframe$ frontiers.

\subsection*{E.1 Turbine configurations}

\begin{table*}[h]
\centering
\caption{Wind farms: specification-based (capacity-weighted) vs frontier-fitted cut-in and rated wind speeds. Specification values $v_c^{\mathrm{spec}}, v_r^{\mathrm{spec}}, v_{co}^{\mathrm{spec}}$ are computed from manufacturer data; fitted values $(v_c^{\mathrm{fit}}, v_r^{\mathrm{fit}})$ are obtained by least-squares on the learned $\theframe$ frontiers, with and without turbine thresholds as inputs. Fitted parameters are obtained post hoc from the learned frontiers and are not directly constrained during training.}
\label{tab:wf-specs-vs-fit}

\begin{tabular}{lcccccccc}
\hline
Farm & \multicolumn{3}{c}{Specification (capacity-weighted)} & \multicolumn{2}{c}{$\theframe$ frontier (no thresholds)} & \multicolumn{2}{c}{$\theframe$ frontier (with thresholds)} \\
\cline{2-4} \cline{5-6} \cline{7-8} \\
Site & $v_c^{\mathrm{spec}}$ & $v_r^{\mathrm{spec}}$ & $v_{co}^{\mathrm{spec}}$ & $v_c^{\mathrm{fit}}$ & $v_r^{\mathrm{fit}}$ & $v_c^{\mathrm{fit}}$ & $v_r^{\mathrm{fit}}$ \\
     & (m/s)                   & (m/s)                   & (m/s)                     & (m/s)                  & (m/s)                  & (m/s)                  & (m/s) \\ \hline
1    & 3.0   & 10.1  & 22.7 & 0.9 & 15.1 & 0.7 & 15.1 \\
2    & 2.5   & 10.5  & 20.0 & 0.5 & 12.2 & 0.5 & 11.6 \\
3    & 3.0   & 10.4  & 25.0 & 0.5 & 15.7 & 0.5 & 16.3 \\
4    & 3.0   & 9.5   & 20.0 & 0.5 & 15.8 & 1.3 & 12.2 \\
5    & 3.0   & 9.0   & 20.0 & 1.4 & 14.4 & 1.6 & 13.1 \\
6    & 3.0   & 13.0  & 25.0 & 1.8 & 18.0 & 1.2 & 18.5 \\ \hline
\end{tabular}

\end{table*}

Table~\ref{tab:wind-farm-specs} reports the nominal capacity, turbine models and key specification parameters for each wind farm that are used as engineering reference points in Section 5.4. For each configuration, we list nominal turbine capacity, hub height, rotor diameter, the number of turbines and the manufacturer-reported cut-in, rated and cut-out wind speeds, together with the corresponding swept rotor area and power-density indicators.

\begin{table*}[h]
\centering
\begin{center}
    \begin{small}
      \begin{sc}
\caption{Technical specifications of turbine configurations at each wind farm. Cut-in, rated and cut-out speeds are manufacturer-reported thresholds for individual turbine models.}
\label{tab:wind-farm-specs}

\begin{tabular}{l l l r r r r r r r r r r}
\hline
Name  & N Cap. & Model & Cap. & Ht     & Rotor   & \#   & Cut-in & Cut-out & RatedSpeed  & Swt.Ar   & PD1      & PD2     \\
      &  &  &          &   & &&        &         &       &    & &  \\
           & (MW)    &          &   (kW)   & (m)     & (m)     &         & (m/s)  & (m/s)   &(m/s)        & (m$^2$) & (W/m$^2$)  & (m$^2$/kW) \\
\hline
F1 & 75   & GW1500/87       & 1500  & 85.0  & 87.0  & 50  & 3.0 & 22 & 9.9  & 5890   & 254.7  & 3.9 \\
F1 & 24   & H93 L-2.0MW     & 2000  & 85.5  & 93.0  & 12  & 3.0 & 25 & 10.8 & 6792.9 & 294.4  & 3.4 \\
F2 & 200  & GW3000/140      & 3000  & 120.0 & 140.0 & 67  & 2.5 & 20 & 10.5 & 15747  & 193.9  & 5.2 \\
F3 & 49.5 & UP86-1500       & 1500  & 80.0  & 86.0  & 33  & 3.0 & 25 & 10.0 & 5809   & 258.2  & 3.9 \\
F3 & 49.5 & UP82-1500       & 1500  & 80.0  & 82.0  & 33  & 3.0 & 25 & 10.8 & 5384   & 278.6  & 3.6 \\
F4 & 30   & FD89A-1500      & 1500  & 85.0  & 89.0  & 20  & 3.0 & 20 & 10.0 & 6221   & 241.1  & 4.1 \\
F4 & 36   & FD116A-2000     & 2000  & 90.0  & 116.0 & 18  & 3.0 & 20 & 9.0  & 10568  & 189.3  & 5.3 \\
F5 & 36   & FD116A-2000     & 2000  & 90.0  & 116.0 & 18  & 3.0 & 20 & 9.0  & 10568  & 189.3  & 5.3 \\
F6 & 96   & XE72            & 2000  & 65.0  & 70.7  & 48  & 3.0 & 25 & 13.0 & 3920   & 510.2  & 2.0 \\
\hline
\end{tabular}
  \end{sc}
    \end{small}
      \end{center}
\end{table*}

\subsection*{E.2 Capacity-weighted specification parameters}

Several farms contain multiple turbine configurations. For each farm, we therefore construct capacity-weighted average specification parameters
$$
    v_c^{\mathrm{spec}}, \quad v_r^{\mathrm{spec}}, \quad v_{co}^{\mathrm{spec}}
$$
by aggregating the cut-in, rated and cut-out speeds across turbine models using installed capacity as weights. Concretely, if farm $s$ comprises configurations indexed by $m$ with turbine capacities $P_{s,m}$ and specification thresholds $v_{c,s,m}$, $v_{r,s,m}$ and $v_{co,s,m}$, then
$$
    v_{c,s}^{\mathrm{spec}}
    =
    \frac{\sum_m P_{s,m}\, v_{c,s,m}}{\sum_m P_{s,m}},
    \quad
    v_{r,s}^{\mathrm{spec}}
    =
    \frac{\sum_m P_{s,m}\, v_{r,s,m}}{\sum_m P_{s,m}},
    \quad
    v_{co,s}^{\mathrm{spec}}
    =
    \frac{\sum_m P_{s,m}\, v_{co,s,m}}{\sum_m P_{s,m}}.
$$
These capacity-weighted values summarise the specification thresholds at the farm level and are used as engineering reference points when comparing to the learned $\theframe$ frontiers. They are not used as hard constraints in training, except in the experiments that explicitly include cut-in, rated and cut-out speeds as additional input features.

Table~\ref{tab:wf-specs-vs-fit} in the main text reports the resulting capacity-weighted $v_c^{\mathrm{spec}}$, $v_r^{\mathrm{spec}}$ and $v_{co}^{\mathrm{spec}}$ for each farm, alongside the corresponding parameters fitted to the frontier capacity-factor curves. The close agreement between specification-based and frontier-fitted rated speeds, and the alignment of effective cut-out speeds with early curtailment behaviour, support the interpretation of the learned $\theframe$ frontiers as physically plausible power curves at the farm level. These capacity‑weighted specification thresholds are not enforced as hard constraints during training; they are used solely for ex post comparison with the learnt frontiers.

\subsection*{E.3 Additional power-curve plots}

\begin{figure*}[ht]
    \centering
     \includegraphics[width=0.32\textwidth]{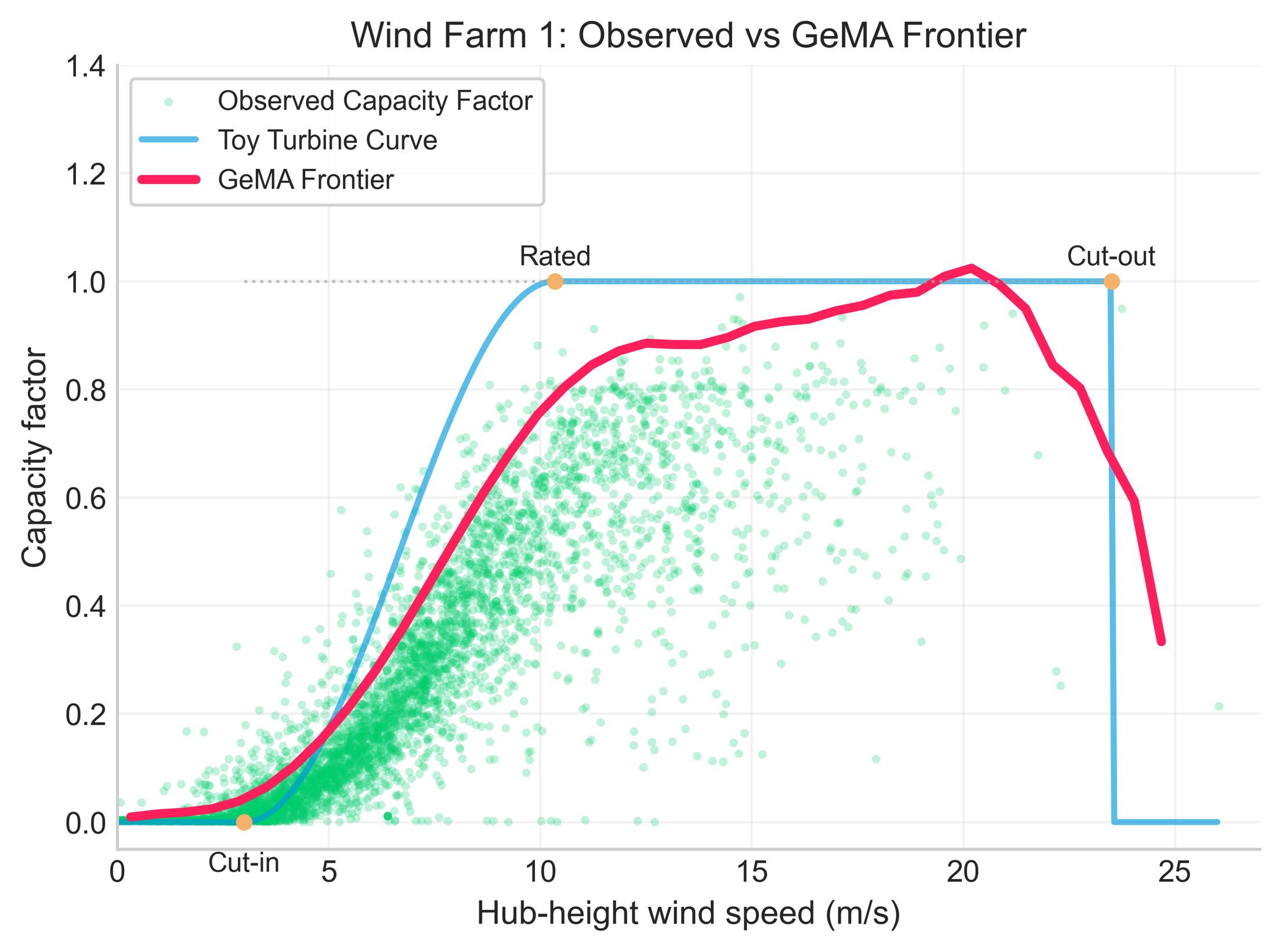}
     \includegraphics[width=0.32\textwidth]{Paper_1_8_GeMA/figures/wf_powercurve/wf2.png}
     \includegraphics[width=0.32\textwidth]{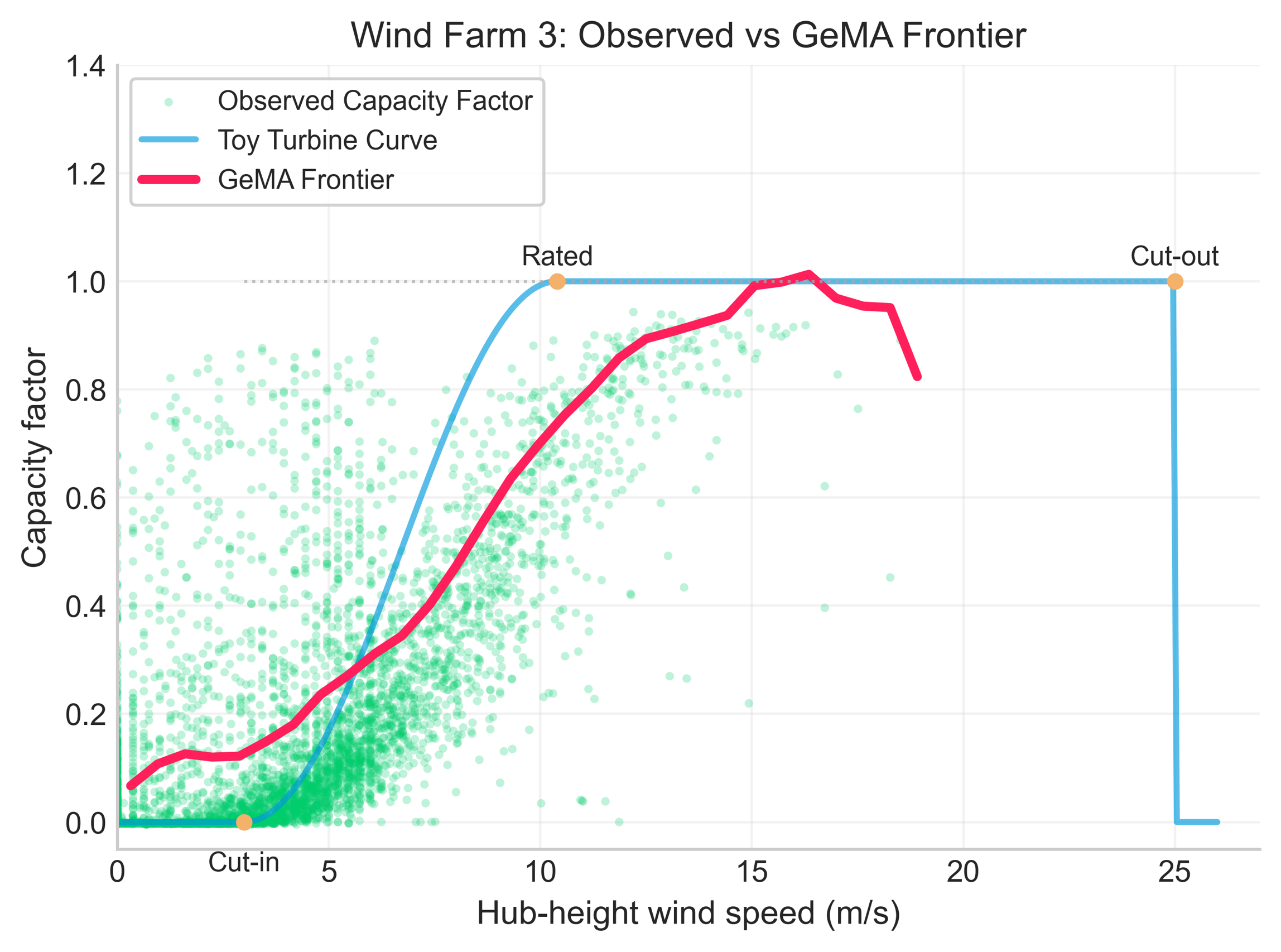}\\[0.5em]
    \includegraphics[width=0.32\textwidth]{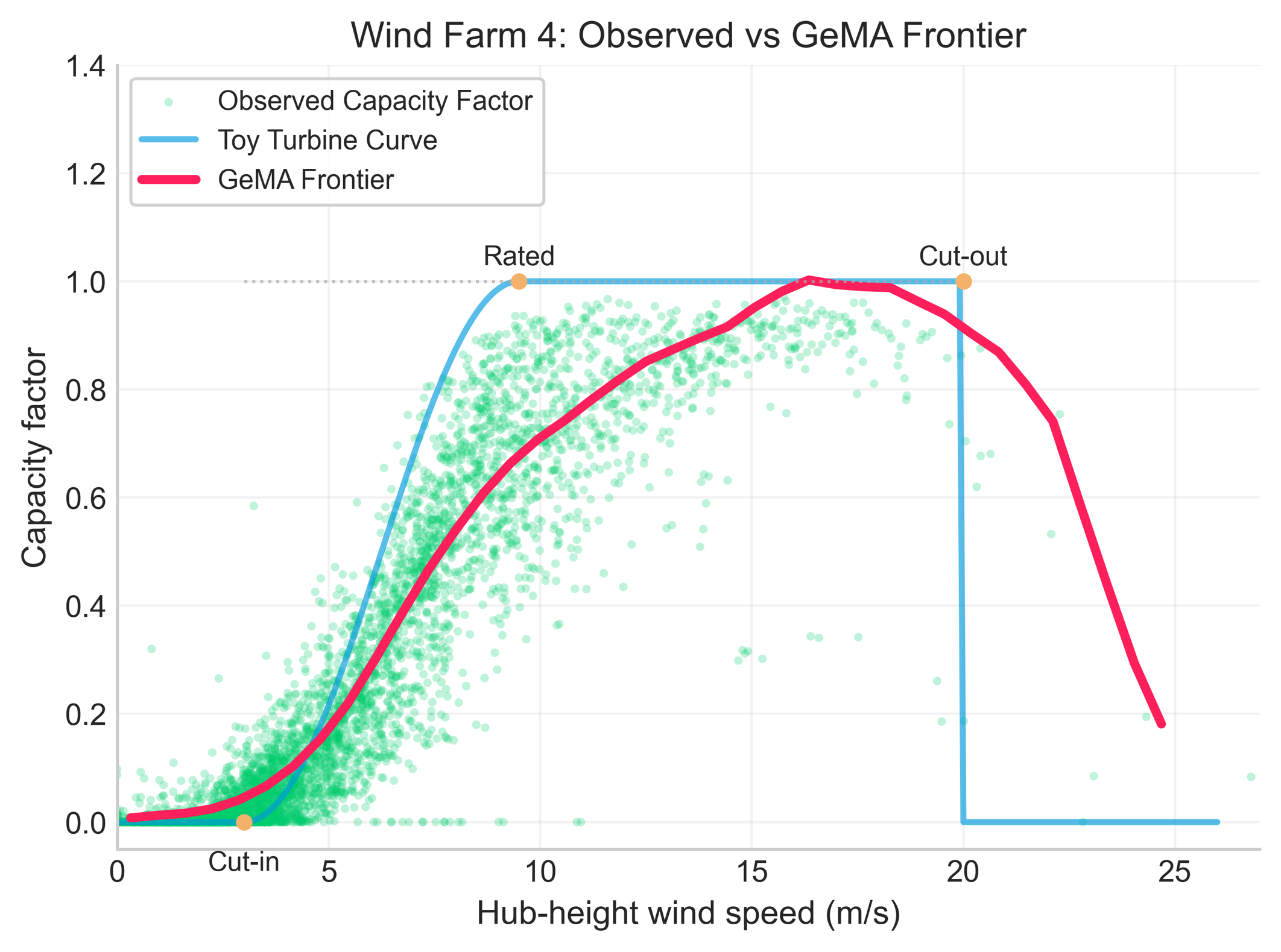}
     \includegraphics[width=0.32\textwidth]{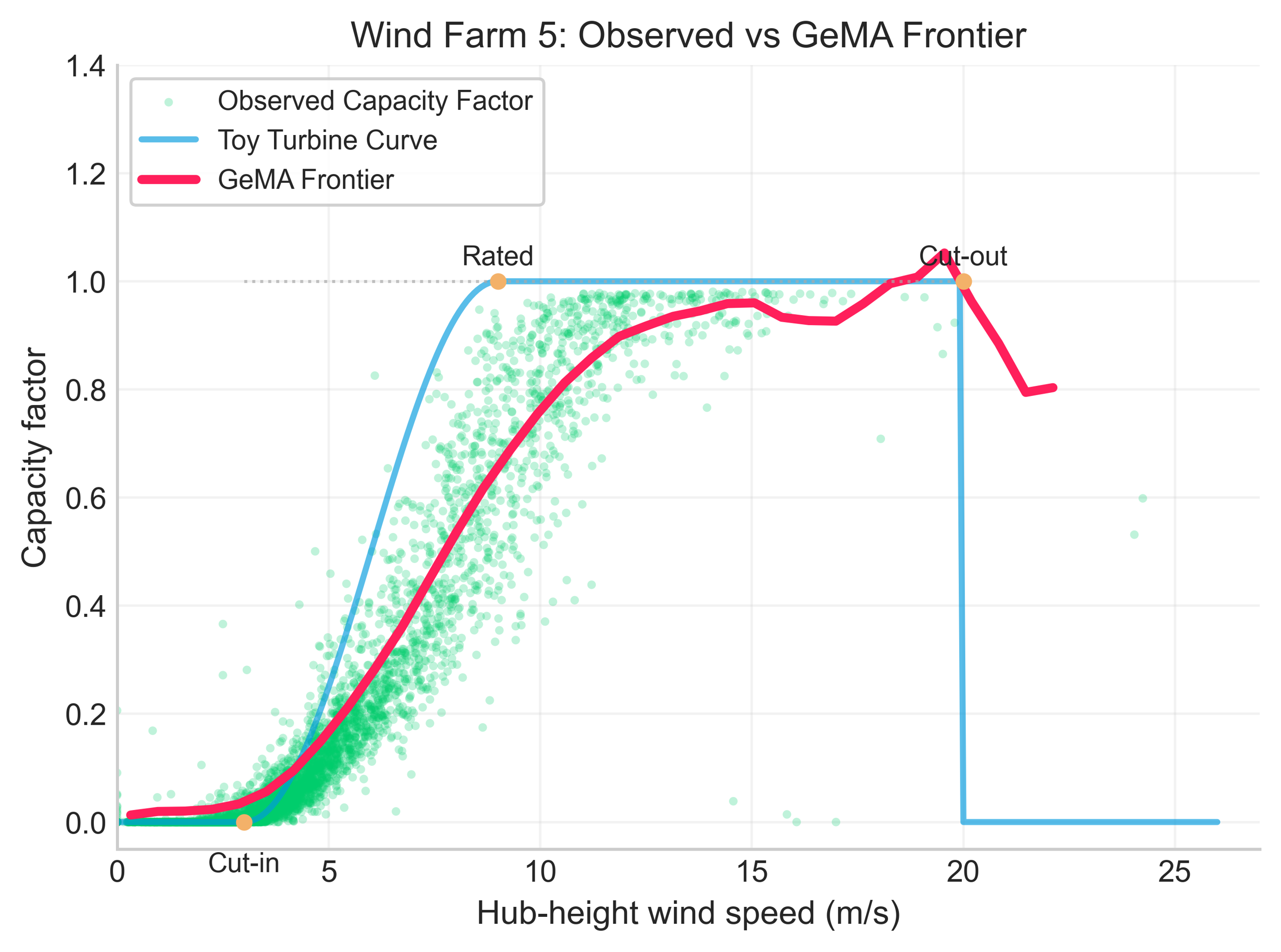}
     \includegraphics[width=0.32\textwidth]{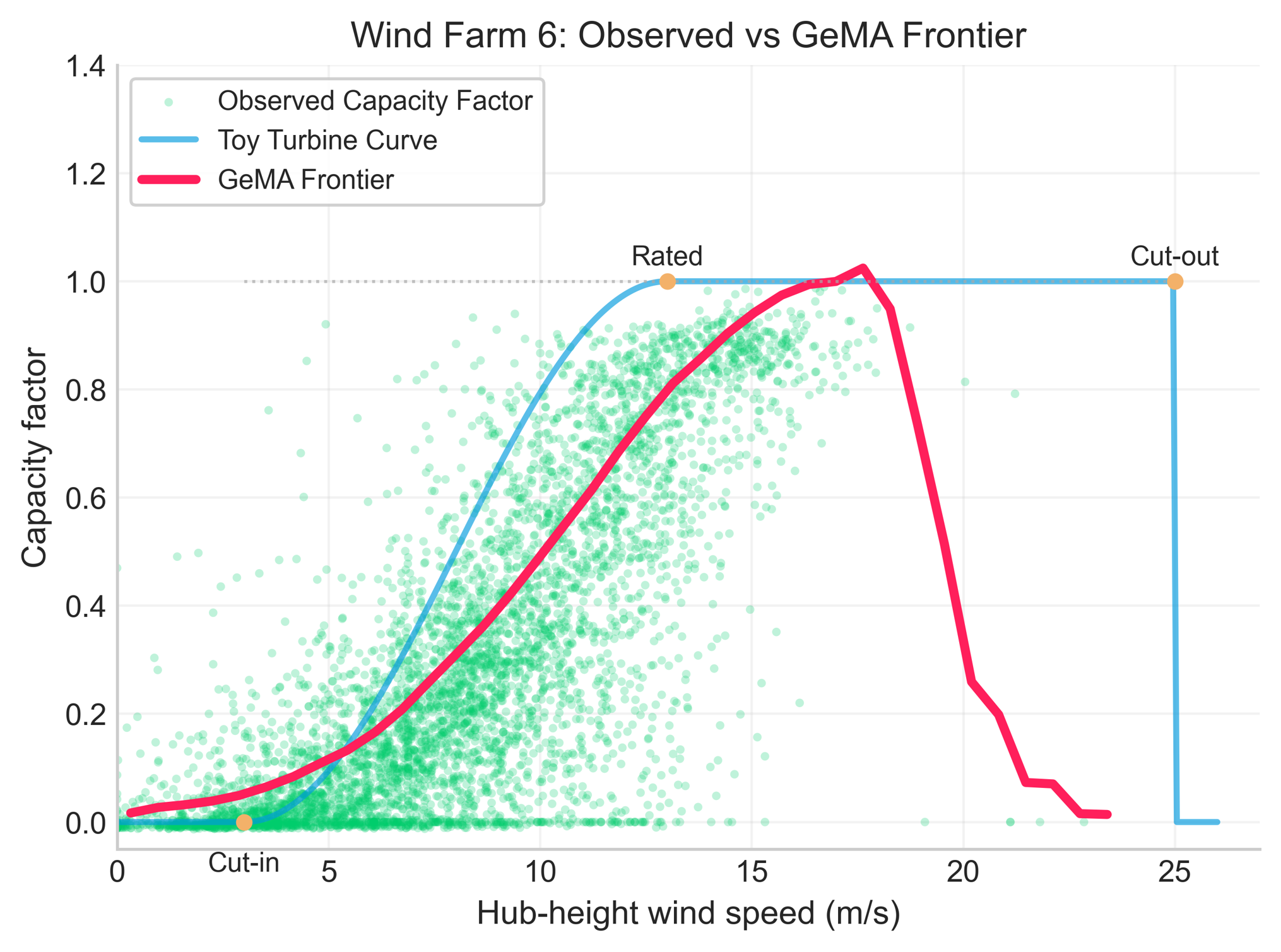}
    \caption{Additional wind farm power-curve plots. For each farm, scatter points show observed capacity factor versus hub-height wind speed; red curves show the learned $\theframe$ frontier (normalised by its 95th percentile); blue curves show specification-based ``toy'' turbine curves constructed from capacity-weighted cut-in, rated and cut-out speeds; orange markers highlight the specification-based operating points. The overall shape and operating points are consistent across farms, with high-wind declines indicating early curtailment where frequent high-wind events occur.}
    \label{fig:wf-power-curves-all}
\end{figure*}

Figure~\ref{fig:wf-power-curves-all} provides additional visualisations of the learnt $\theframe$ frontiers for all six wind farms. For each farm, we plot observed capacity factor against hub-height wind speed, overlaid with the corresponding learnt frontier and a specification-based ``toy'' turbine curve constructed from capacity-weighted cut-in, rated, and cut-out speeds. As in Section~\ref{subsec:wind}, frontier capacity factors are normalised by their 95th percentile per farm so that the empirical plateau aligns approximately with $\mathrm{CF}=1$. All plots normalise capacity factors by each farm’s 95th percentile, so that the plateau of the empirical frontier is approximately aligned at unity for visual comparison.

Across all farms, the learned frontiers reproduce the characteristic three-stage shape of turbine power curves. The onset of generation occurs near the specification-based cut-in speeds, the non-linear ramp-up region aligns with the mid-range of wind speeds, and the plateau lies close to the nominal rated capacity factor. In farms with frequent high-wind event,s the frontier exhibits a decline in capacity factor at high wind speeds, often somewhat below the specification-based cut-out speeds, consistent with early curtailment and grid or turbine protection. For farms with limited high-wind data, the frontier remains effectively flat in the upper range and we do not attempt to infer an effective cut-out point.

\section*{Appendix F: Additional COMET results}
\label{app:comet_extra}

This appendix reports additional diagnostics for the COMET case study, complementing Section 5.1

\subsection*{F.1a Cluster interpretation using observable characteristics}

Table~\ref{tab:app_comet_cluster_observables} reports cluster-wise medians (and IQRs) of key observed variables (route length, stations, capacity, passenger-km, car-km, staff), providing an interpretable description of each latent peer group.

\begin{table}[ht]
\centering
\caption{COMET (test period 2018--2019): cluster-wise summaries of observable characteristics. Reported as median (interquartile range) over operator--year observations.}
\label{tab:app_comet_cluster_observables}

\begin{tabular}{lccccccc}
\hline
Cluster & \# Test obs. &Route length &Stations &Capacity &Staff &Passenger-km &Car-km \\
\hline
A & 27 &120.7 (231.1) &100.5 (120.5) &40.0 (16.7) &3818 (7980) &3107 (5184) &94.4 (172.0) \\

B & 28 &186.4 (266.1) &87.0 (190.3) &43.2 (15.3) &5688 (11720) &5532 (13115) &140.1 (281.8) \\

C & 16 &141.2 (135.0) &104.0 (88.3) &40.5 (10.4) &5600 (6703) &6660 (14090) &128.9 (211.5) \\

D & 13 &115.1 (116.9) &91.0 (59.0) &44.4 (10.8) &4529 (11844) &6201 (7541) &114.7 (164.3) \\
\hline
\end{tabular}

\end{table}

\subsection*{F.1b Cluster-wise baseline score distributions}

Table~\ref{tab:app_comet_cluster_baselines} reports cluster-wise medians and IQRs of key observed variables and baseline scores, providing an interpretable description of each latent peer group.

\begin{table}[ht]
\centering
\caption{COMET (test period 2018--2019): cluster-wise summaries of observable scale and baseline scores. Reported as median (IQR) over operator--year observations.}
\label{tab:app_comet_cluster_baselines}

\begin{tabular}{lccccc}
\hline
Cluster & \# Test obs. & Route length (IQR) & DEA $\hat\theta$ (IQR) &$\theframe$ score (IQR) & RF proxy $u$ (IQR) \\
\hline
A & 27 & 120.7 (231.1) & 1.000 (0.050) & -0.255 (0.319) & 0.110 (0.283) \\
B & 28 & 186.4 (266.1) & 0.996 (0.073) & -0.447 (0.307) & 0.075 (0.269) \\
C & 16 & 141.2 (135.0) & 0.995 (0.101) & -0.310 (0.505) & 0.043 (0.185) \\
D & 13 & 115.1 (116.9) & 1.000 (0.043) & -0.555 (0.371) & 0.005 (0.179) \\
\hline
\end{tabular}

\end{table}

\subsection*{F.2 Posterior cluster assignment confidence}
We report summary statistics of GMM posterior probabilities $\max_c p_{ic}$ in Table~\ref{tab:app_comet_cluster_posterior} to quantify assignment certainty.

\begin{table}[ht]
\centering
\caption{COMET (test period 2018--2019): posterior cluster assignment confidence from the GMM on latent technology space. Reported as median (IQR) of $\max_c p_{ic}$ and the share of observations exceeding common confidence thresholds.}
\label{tab:app_comet_cluster_posterior}

\begin{tabular}{lcccc}
\hline
Cluster & \# Test obs. & $\max_c p_{ic}$ (IQR) & Share $\max_c p_{ic}\ge 0.8$ & Share $\max_c p_{ic}\ge 0.9$ \\
\hline
A & 27 & 0.534 (0.195) & 11.1\% & 0.0\% \\
B & 28 & 0.611 (0.305) & 21.4\% & 7.1\% \\
C & 16 & 0.759 (0.346) & 43.8\% & 18.8\% \\
D & 13 & 0.797 (0.272) & 46.2\% & 23.1\% \\
\hline
\end{tabular}

\end{table}

These values suggest that clusters C and D are more compact in latent space, while clusters A and B exhibit softer boundaries, consistent with greater internal heterogeneity. 

\subsection*{F.3 Robustness/certification diagnostics}
We report in Table~\ref{tab:app_comet_rc_percentiles} for certification radius percentiles and coverage statistics to characterise where the model’s local robustness guarantees are strongest.

\begin{table}[ht]
\centering
\caption{COMET: certification radius percentiles (whitened-input robustness diagnostic).}
\label{tab:app_comet_rc_percentiles}
\begin{tabular}{lc}
\hline
Percentile & Certification radius \\
\hline
0\%  & 0.135 \\
5\%  & 0.177 \\
25\% & 0.249 \\
50\% & 0.322 \\
75\% & 0.364 \\
95\% & 0.440 \\
99\% & 0.475 \\
\hline
\end{tabular}
\end{table}

The distribution of certification radii indicates that local robustness guarantees vary across systems, with higher radii concentrated among centrally located latent representations


\end{document}